\documentclass[11pt]{article}

\usepackage{fullpage}
\usepackage[english]{babel}

\usepackage[utf8]{inputenc} % allow utf-8 input
\usepackage[T1]{fontenc}    % use 8-bit T1 fonts
\usepackage{hyperref}       % hyperlinks
\usepackage{csquotes}
\hypersetup{
colorlinks,
linkcolor={red!70!black},
citecolor={blue!70!black},
urlcolor={blue!80!black}
}
\usepackage{url}            % simple URL typesetting
\usepackage{booktabs}       % professional-quality tables
\usepackage{amsfonts}       % blackboard math symbols
\usepackage{nicefrac}       % compact symbols for 1/2, etc.
\usepackage{microtype}      % microtypography
\usepackage{setspace}
\usepackage{caption}
\usepackage{subcaption}
\usepackage{amsmath,amssymb,amsthm,bbm}

\usepackage{cleveref}
\creflabelformat{equation}{#2#1#3}
\usepackage{amsmath}
\usepackage{mathtools}
\usepackage[dvipsnames]{xcolor}
\usepackage{algorithm}
\usepackage{algorithmic}
\usepackage{xspace}

\usepackage{enumerate}\usepackage[inline]{enumitem}
\usepackage{etoolbox,siunitx}
\robustify\bfseries
\usepackage{multirow,makecell}
\usepackage{threeparttable}
 
\usepackage[
 backend=biber,
 style=numeric,
 giveninits=true,
 maxcitenames = 2,
 maxbibnames = 10,
 ]{biblatex}
 \DeclareSourcemap{
  \maps[datatype=bibtex]{
    \map{
      \step[fieldset=issn, null]
      \step[fieldset=doi, null]
      \step[fieldset=url, null]
    }
    \map[overwrite=true]{
      \step[fieldsource=fjournal]
      \step[fieldset=journal, origfieldval]
    }
  }
}
\usepackage{graphicx}

\theoremstyle{plain}
\newtheorem{theorem}{Theorem}[section]
\newtheorem{proposition}[theorem]{Proposition}
\newtheorem{lemma}[theorem]{Lemma}
\newtheorem{corollary}[theorem]{Corollary}
\theoremstyle{definition}

\theoremstyle{remark}

\renewcommand{\epsilon}{\varepsilon}

\newcommand*{\MyDef}{\mathrm{\tiny def}}
\newcommand*{\eqdefU}{\ensuremath{\mathop{\overset{\MyDef}{=}}}}\renewcommand*{\triangleq}{\eqdefU}

\def\:#1{\protect \ifmmode {\mathbf{#1}} \else {\textbf{#1}} \fi}

\newcommand{\wt}[1]{\widetilde{#1}}

\newcommand{\bx}{\mathbf{x}}
\newcommand{\by}{\mathbf{y}}

\newcommand{\bA}{\mathbf{A}}
\newcommand{\bB}{\mathbf{B}}

\newcommand{\bI}{\mathbf{I}}

\newcommand{\bK}{\mathbf{K}}

\newcommand{\bV}{\mathbf{V}}
\newcommand{\bW}{\mathbf{W}}
\newcommand{\bX}{\mathbf{X}}

\DeclareMathOperator*{\argmax}{arg\,max}
\DeclareMathOperator*{\logdet}{log\,det}

\renewcommand{\epsilon}{\varepsilon}
\newcommand{\bigotime}{\mathcal{O}}
\newcommand{\abigotime}{\wt{\mathcal{O}}}

\DeclareMathOperator*{\polylog}{polylog}

\newcommand{\normsmall}[1]{\Vert #1 \Vert}

\newcommand{\transp}{\mathsf{\scriptscriptstyle T}}

\newcommand{\Real}{\mathbb{R}}

\newcommand{\rkhs}{\mathcal{H}}

\newcommand{\kerfunc}{\mathrm{k}}

\newcommand{\ie}{i.e.,\xspace}
\newcommand{\eg}{e.g.,\xspace}

\newcommand{\armset}{\mathcal{A}}

\newcommand{\fnorm}{F}
\newcommand{\Narm}{A}

\newcommand{\fb}[1]{\normalfont \texttt{fb}(#1)}

\newcommand{\bkb}{\textsc{BKB}\xspace}
\newcommand{\bbkb}{\textsc{BBKB}\xspace}

\newcommand{\gpopt}{GP-Opt\xspace}
\newcommand{\uq}{q}
\newcommand{\gpucb}{\textsc{GP-UCB}\xspace}
\newcommand{\bgpucb}{\textsc{GP-BUCB}\xspace}
\newcommand{\bgpts}{\textsc{GP-BTS}\xspace}

\newcommand{\mugpucb}{\textsc{mini}-\textsc{GP-UCB}\xspace}
\newcommand{\mugpei}{\textsc{mini}-\textsc{GP-EI}\xspace}
\newcommand{\mumeta}{\textsc{mini}-\textsc{META}\xspace}

\newcommand{\rsoful}{\textsc{RS-OFUL}\xspace}
\newcommand{\gpei}{\textsc{GP-EI}\xspace}
\newcommand{\gpts}{\textsc{GP-TS}\xspace}

\newcommand{\epoch}{h}
\newcommand{\numepoch}{h}

\title{\bf Scaling Gaussian Process Optimization \\ by Evaluating a Few Unique Candidates Multiple Times}

\date{}

\makeatletter
\newcommand{\printfnsymbol}[1]{%
  \textsuperscript{\@fnsymbol{#1}}%
}
\makeatother

\author{%
  {\bf Daniele Calandriello} \hfill \hspace{12.5em}{\small\texttt{dcalandriello@deepmind.com}}\\
  {\small \it DeepMind Paris, France \hfill \hspace{12em}}\\   
  \and
{\bf Luigi Carratino}\hfill \hspace{13em}\small \texttt{luigi.carratino@dibris.unige.it}\\
  {\small \it MaLGa - DIBRIS, University of Genova, Italy \hfill \hspace{12em}}\\
  \and
 {\bf Alessandro Lazaric} \hfill \hspace{18.5em}{\small\texttt{lazaric@fb.com}}\\
 {\small \it Facebook AI Research Paris, France \hfill \hspace{12em}}\\ 
  \and
   {\bf Michal Valko} \hfill \hspace{19em}{\small\texttt{valkom@deepmind.com}}\\
 {\small \it DeepMind Paris, France \hfill \hspace{12em}}\\ 
  \and
  {\bf Lorenzo Rosasco} \hfill \hspace{15em}{\small\texttt{lorenzo.rosasco@unige.it}}\\
  {\small \it MaLGa - DIBRIS, University of Genova, Italy  \hfill \hspace{12em}}\\
  {\small \it Istituto Italiano di Tecnologia, Genova, Italy \hfill \hspace{12em}}\\
  {\small \it CBMM - MIT, Cambridge, MA, USA \hfill \hspace{12em}}
}

\begin{document}

\maketitle

\begin{abstract}
Computing a Gaussian process (GP) posterior has a computational cost cubical in the number of historical
points. A reformulation of the same GP posterior
highlights that this complexity mainly depends on how many \emph{unique}
historical points are considered.
This can have important implication in active learning settings, where the set
of historical points is constructed sequentially by the learner. We show that sequential black-box optimization based on GPs
(GP-Opt) can be made efficient by sticking to a candidate solution for multiple
evaluation steps and switch only when necessary. 
Limiting the number of
switches also limits the number of unique points in the history of the GP. Thus, the efficient GP reformulation can be used to exactly and cheaply compute
the posteriors required to run the GP-Opt algorithms.
This approach is especially useful in real-world
applications of GP-Opt with high switch costs (e.g. switching
chemicals in wet labs, data/model loading in hyperparameter optimization).
As examples of this meta-approach, we modify two
well-established GP-Opt algorithms, GP-UCB and GP-EI, to switch candidates as
infrequently as possible adapting rules from batched GP-Opt.
These versions preserve all the theoretical no-regret
guarantees while improving
practical aspects of the algorithms such as runtime, memory complexity, and the
ability of batching candidates and evaluating them in parallel.
\end{abstract}
\section{Introduction}

Bayesian and bandit optimization \cite{mockus1989bayesian, lattimore2020bandit}
 are two well established frameworks for black-box
function optimization under uncertainty. They model the
optimization process as a sequential learning problem. For an unknown
function $f$ and a decision set $\armset$ (e.g., $\armset \subseteq \Real^d$) at each step $t$ the learner:
\\
(1) selects candidate $\bx_t \in \armset$  by maximizing an
acquisition function $u_t$ as a surrogate of $f$;\\
(2) receives a noisy feedback
\footnote{Actions or arms are also used for $\bx_t$; and rewards for $y_t$.}
$y_t \triangleq f(\bx_t) + \eta_t$ from the environment, where $\eta_t \stackrel{\text{i.i.d.}}{\sim} \mathcal{N}(0,\xi^2)$;\\[0.3em]
(3) uses $y_t$ to improve $u_t$'s approximation of $f$.

A common approach to measure the learner's convergence
is to study the \textit{cumulative regret}
$R_T \triangleq \sum_{t=1}^T f^{\star} - f(\bx_t)$
of the learner's choice of candidates compared to the optimum of
the function $f^\star = \max_{\bx} f(\bx)$. A learner is deemed
no-regret when it converges on average (i.e., $\lim_{T \rightarrow \infty}
R_T/T \rightarrow 0$). Thanks to their flexibility
and capability of modeling uncertainty, Gaussian processes (GP)
\cite{rasmussen_gaussian_2006} have emerged as an effective choice for regret minimization, jump-starting the field of GP optimization (\gpopt).
Decision rules that leverage GPs to estimate upper confidence bounds (\gpucb
\cite{srinivas2010gaussian, chowdhury2017kernelized}) or expected
improvement (\gpei
\cite{wang2014theoretical}) provably achieve a regret\footnote{The notation $\abigotime$ ignores $\polylog$ terms.}
 $\abigotime(\gamma_T\sqrt{T})$ over $T$ steps. Here,~$\gamma_T$
is the so-called maximum information gain associated with $\armset$
and the GP. It captures in a data-adaptive way the implicit number of parameters
of the non-parametric GP model, quantifying the
effective complexity of the optimization problem.
Under appropriate assumptions $\gamma_T$ grows slowly
in $T$ making GP based methods no-regret
\cite{srinivas2010gaussian,scarlett2017lower}.
However, as most non-parametric approaches, \gpopt algorithms
face scalability issues when $T$ grows large, severely limiting
their applicability to large scale settings.
For example, in some 
hyper-parameter optimization problems
the solutions can still improve even after tens of thousands of
candidate evaluations (e.g. the NAS-bench search \cite{ying2019bench} evaluated 5 million candidates).
Even more, for recommender systems 
that continuously adapt to the users' behaviour, the horizon $T$ is unbounded
and the optimization process never ends (e.g. \cite{saito2020open,li2010contextual} record tens of millions of interactions with users).
Many approaches have been proposed to extend \gpopt algorithms
to these settings.

\subsection{Approaches to scalable no-regret GP optimization}
%Building on this simple sequential optimization protocol, many GP optimization
%methods have been developed to address several scalability bottlenecks present
%in the original formulation, while at the same time preserving convergence and
%no-regret guarantees.

GP's most commonly considered limitation is the \textbf{computational
bottleneck}, which stems from the high cost of quantifying uncertainty using
exact inference on the GP model. In particular, in the traditional
i.i.d.~learning setting the complexity of computing a GP posterior over $n$
training points is usually quantified as
$\abigotime(n^3)$ time and $\abigotime(n^2)$ space \cite{rasmussen_gaussian_2006},
which makes training GPs on large dataset unfeasible in general.
%This complexity analysis mostly carries over to the GP optimization settings,
%where we must compute a new posterior after every new candidate selection but
%can use incremental computation to update the GP posterior in $\abigotime(t^2)$
%time, giving us a baseline complexity of $\abigotime(\sum_{t=1}^T t^2) = \abigotime(T^3)$ and space complexity of $\abigotime(T^2)$.
%Once again this lack of scalability would relegate GPs only to short
%optimization processes, and over the years multiple methods emerged to resolve
%this computational bottleneck.
A similar complexity analysis in the \gpopt settings
replaces $n$ with the $T$ candidates chosen by the algorithm,
giving us a baseline complexity of $\abigotime(T^3)$ time 
and $\abigotime(T^2)$ space complexity.
Two main approaches have emerged to reduce this complexity.
The most common can provably preserve no-regret guarantees, and is based on
approximating the GP model, using either inducing points
\cite{quinonero-candela_approximation_2007,calandriello_2019_coltgpucb} or
quadrature Fourier features \cite{mutny_efficient_2018}.
Another approach
is to use approximate posterior inference, e.g., using iterative solvers \cite{gardner2018gpytorch}, but without no-regret guarantees.

Beyond computations, vanilla \gpopt algorithms suffer from a
\textbf{sequential experimentation bottleneck}, which stems from having to wait
for feedback at every step of the sequential interaction protocol.
Whenever each function evaluation is time consuming this bottleneck can actually dominate the computational one, and even computationally efficient
\gpopt algorithms cannot scale as they spend most of their time waiting for feedback.
However, even in this slow evaluation setting scalability can be  achieved by selecting batches of candidates at the same time and evaluating them in parallel.
Experimental parallelism greatly reduces total evaluation time, and millions of evaluations are possible resulting in much better final performance \cite{ying2019bench}.
The key trade-off is between batch sizes and regret, with larger batches increasing the potential for parallelism but also increase feedback delay and regret.
Several approaches have been proposed over the year to manage this trade-off \cite{desautels_parallelizing_2014,kathuria_batched_2016,daxberger17a}.

Overall, to match existing state of the art \gpopt algorithms
in guarantees for 
regret, computational cost and batching capability,
an algorithm should achieve
$\abigotime(\gamma_T\sqrt{T})$ regret with a $\abigotime(T\gamma_T^3)$ time and
$\abigotime(T\gamma_T + \gamma_T^2)$ space complexity, and only $\abigotime(\gamma_T)$ rounds
of interactions (i.e., batches).

\subsection{Contributions}
In this paper we propose a meta-algorithm, \mumeta, that we use to generate scalable versions of many popular \gpopt algorithm. 
Applying \mumeta to \gpucb
and \gpei we obtain two new \textsc{mini}-variants, \mugpucb and \mugpei, that
achieve performance comparable to existing state-of-the-art scalable \gpopt algorithms, but
rely on very different underlying principles to achieve scalability.

There are two main building blocks in \mumeta. The first
is a reformulation of the posterior of a GP
that highlights that scalability in a GP does not depend on
the number of points in the GP, but rather on the number of \emph{unique}
(i.e., different from all others) points.
Indicating with $\uq_t$ the number of unique points up to time $t$, the time complexity
of computing a GP can then be reduced from $\abigotime(T^3)$ to $\abigotime(\uq_T^3)$.
This is an elementary property of GPs, but its applicability
greatly depends on the underlying task. It has little impact and is not used in i.i.d.\,learning settings
where duplicates in the
training data are exceedingly rare.
It can bring more benefits in settings were
multiple evaluations of the same candidate
are naturally part of the solution,
and it has been leveraged
when possible in stochastic optimization \cite{picheny2013quantile} and active learning \cite{binois2019replication}.
In this paper we show that this approach can be taken a step further in \gpopt tasks. In particular, the second building block of \mumeta is a meta-strategy that given a \gpopt algorithm as input, modifies the candidate selection strategy of the input learner to explicitly
minimize $\uq_t$ to improve performance while preserving the same regret rate guarantees as the original candidate selection strategy.

Out of the many meta-strategies that one can consider, a simple
yet effective approach that we advocate is to limit the number of times
the candidate selection strategy is allowed to switch candidates,
since the number of candidate switches clearly bounds $\uq_t$. This can be done
transparently for many input learners by using the selection strategy, and then
sticking to the same candidate for multiple steps before invoking the selection
strategy again.
Furthermore, the rule we derive to decide when to switch candidate
does not depend on the function feedback.
Therefore the algorithm can first select a candidate,
then choose how many steps it should stick to it, and finally evaluate
the candidate multiple times in parallel. In other words,
our meta-approach can not only retro-fit existing \gpopt
algorithms to be computationally scalable, but also experimentally
scalable thanks to batching.

Theoretically, we study the impact of this generic modification
both under the lenses
of model freezing in the linear bandits~\cite{abbasi2011improved} and
delayed feedback in batch \gpopt
\cite{desautels_parallelizing_2014,calandriello2020near}.
Using these tools we derive strong guarantees for two new variants of \gpucb
and \gpei that balance between minimizing unique actions and achieving low
regret, hence dubbed minimize-unique-\gpucb (\mugpucb) and
minimize-unique-\gpei (\mugpei). Thanks to a careful balancing of the
trade-off, both algorithms match the $\abigotime(\gamma_T\sqrt{T})$ regret of
their traditional counterparts, while evaluating a number of unique candidates
$\uq_t \leq \abigotime(\gamma_t)$ that scales only with the maximum
information gain. Moreover, these algorithms are extremely scalable,
requiring only $\abigotime(T + A\uq_T^3) \leq \abigotime(T + A\gamma_T^3)$ time,
where $A$ is the time required to optimize $\uq_t$,
$\abigotime(T\gamma_T + \gamma_T^2)$ space, and $\abigotime(\gamma_T)$ rounds
of interaction.

Our approach also highlights an even less studied family of scalability bottlenecks for GPs, the \textbf{candidate switch bottlenecks}.
Similar to the sequential experimentation bottleneck, this is an experimental bottleneck that stems from the fact
that in some cases, changing the candidate being evaluated can be time or cost expensive
(\eg even if testing a chemical might be fast, re-calibrating instruments for different chemicals might be expensive).
%Alternatively, one can see this bottleneck as an effect of economy of scale:
%often evaluating multiple times the same candidate can be much
%simpler than evaluating a diverse set of candidates (\eg up-front costs in medical trial
%to manufacture many copies of a single candidate drug vs.~single copies of many candidates).

Finally, our approach is unique in the scalable GP literature in the
sense that it completely side-steps the necessity of approximating
inference or the GP model;
avoiding the need of inducing points, randomized sketching, or iterative solvers.
Instead, algorithms based on
\mumeta can use exact inference on an exact GP
model thanks to the efficient posterior reformulation and the
different optimization path chosen to minimize unique
candidates. This greatly simplifies the algorithm, making it more
broadly applicable.
%As an example of how much this broadens the
%applicability of our approach compared to existing methods, while
%so far most sketching methods have been focused on \gpucb,
%\mumeta can easily be applied to \gpei to obtain the first scalable
%and provably no-regret \gpopt algorithm based on EI.
%Not relying on any kind of GP approximation also makes our meta-approach
%sufficiently general to warrant interest in other potential applications,
%both in \gpopt (e.g., other acquisition functions beyond UCB and EI),
%and in the broader active learning setting.

%$\bx_t$ over multiple
%time steps, effectively introducing copies $\bx_t = \bx_{t+1} = \dots$ of the
%candidate in the history each with its own \emph{different} feedback $y_t \neq
%y_{t+1}$. The computational efficiency then simply come by replacing all the
%multiple copies with a single re-weighted copy with appropriately re-weighted
%feedback.

\section{Preliminaries}\label{sec:preliminaries}

\textbf{Notation.} We denote with $[t]=\{1,\ldots,t\}$ the set of integers up
to $t$. At step $t$, the matrix $\bX_t  \triangleq [\bx_1, \dots,
\bx_t]^\transp \in \Real^{t \times d}$ contains all the candidates selected so
far, and $\by_t \triangleq [y_1, \dots, y_t]^\transp$ their
corresponding feedbacks. We define a batch as a sequence of steps between $t
\leq t'$ such that the learner must select all candidates $\bx_t, \dots
\bx_{t'}$ before observing the corresponding feedback $y_t,\dots,y_{t'}$. To
connect steps and batches we adopt a standard notation 
\cite{desautels_parallelizing_2014,calandriello2020near} where $\fb{t}$
indicates the last step of the previous batch, \ie at step $t$ the learner has
access only to feedback $\by_{\fb{t}}$ up to step $\fb{t}$.

\textbf{Gaussian processes.}
GPs \cite{rasmussen_gaussian_2006} are usually described using a mean function $\mu$, which we assume to be zero, and a (bounded) covariance or kernel function  $\kerfunc: \armset\times\armset\rightarrow [0,\kappa^2]$.
Based on the prior $\mu$ and~$\kerfunc$, the posterior of the GP conditioned on some data $\bX_t, \by_t$ is traditionally formulated as
\begin{align}
\begin{aligned}
{\mu}_t(\bx_i) &= \kerfunc(\bx_i, \bX_t)(\bK_t + \lambda\bI)^{-1}\bK_t\by_t,\\
{\sigma}_t^2(\bx_i) &= \kerfunc(\bx_i, \bx_i) - \kerfunc(\bx_i, \bX_t)(\bK_t + \lambda\bI)^{-1}\kerfunc(\bX_t, \bx_i),
\end{aligned}
\label{eq:post-gp}
\end{align}
where we shortened $\bK_t = \kerfunc(\bX_t, \bX_t)$, and use
only the subscript $t$ to indicate that
${\mu}_t$ and $\sigma_t$ are conditioned on all data up to time $t$.
Note also that $\mu_t$ and $\sigma_t$ are only proper Bayesian
GP posterior when $\lambda = \xi^2$, but we leave here
$\lambda$ as a free parameter as it will be useful in frequentist
regret analysis.
%We also use $GP(\mu_t, \kerfunc_t)$
%to indicate the posterior distribution over functions induced by the GP
%and data $\bX_t$.
Given prior and posterior of the GP, we can also define the information gain
between them as 
\vspace{-.25\baselineskip}
\begin{align*}
\textstyle
\gamma(\bX, \by) \triangleq \tfrac{1}{2}\logdet(\bI + \xi^{-2}\kerfunc(\bX,\bX)).
\end{align*}
Based on $\gamma(\bX,\by)$ we can also define the maximum information gain
$\gamma_t \triangleq \max_{\bX: |\bX| = t} \gamma(\bX, \by)$ at step $t$ as a
worst-case bound on the complexity of the optimization process.
Large amounts of research have been dedicated to characterizing how $\gamma_T$
behaves under different assumptions on $\armset$ and $\kerfunc$ \cite{srinivas2010gaussian,scarlett2017lower}. This includes sufficient conditions for
$\gamma_T$ to be sublinear, both with polynomial and logarithmic rates in $T$.
We leave the reader to recent surveys \cite{pmlr-v130-vakili21a} for more results, and treat here
$\gamma_T$ as a desirable data-adaptive measure of the intrinsic
complexity of the problem.
%\vspace{-.25\baselineskip}
%\begin{align*}
%\textstyle
%\deff(\bX_t) = \tfrac{1}{\lambda}\sum_{s=1}^t \sigma_t^2(\bx_s) = \Tr(\bK_t(\bK_t + \lambda\bI)^{-1}).
    %\end{align*}
%\vspace{-1.75\baselineskip}
%Intuitively, $\deff(\bX_t)$ captures the effective
%number of parameters in $f$, \ie the posterior $f$ can be represented using roughly $\deff(\bX_t)$ coefficients.
%We use $\deff$ to denote $\deff(\bX_T)$ at the end of the process.
%Note that $\deff$ is equivalent
%to the maximal conditional mutual information $\gamma_T$ of the GP
%\cite{srinivas2010gaussian}, up to logarithmic terms~\citespecific{Lem.~1}{calandriello_2017_icmlskons}.

\textbf{Optimistic Gaussian process optimization.}
Given a GP posterior, we can construct an \textit{acquisition function} $u_t(\cdot): \armset \to \Real$
to guide us in the candidate selection.
We focus specifically on acquisition functions based on upper confidence bounds (\gpucb~\cite{srinivas2010gaussian})
%,Thompson sampling,\eg \gpts~\cite{chowdhury2017kernelized}),
and expected improvement (\gpei \cite{mockus1989bayesian})
\begin{align}
&u_t^{\gpucb}(\bx) = \mu_t(\bx) + \beta_t^{\gpucb}\sigma_t(\bx),\label{eq:acq-gpucb}\\
%u_t^{\gpts}(\bx) &\sim GP(\mu_t, \beta_t^{\gpts}\kerfunc_t),\\
&u_t^{\gpei}(\bx) = \beta_t^{\gpei}\sigma_t(\bx)\left[\big(\tfrac{z}{\beta_t^{\gpei}}\big)
\textrm{CDF}_{\mathcal{N}}\big(\tfrac{z}{\beta_t^{\gpei}}\big) + \textrm{PDF}_{\mathcal{N}}\big(\tfrac{z}{\beta_t^{\gpei}}\big)\right]\nonumber,
\end{align}
where 
$z = \tfrac{\mu_t(\bx) - \max_{\bx'}\mu_t(\bx')}{\sigma_t(\bx)}$,
$\textrm{CDF}_{\mathcal{N}}(\cdot)$ and $\textrm{PDF}_{\mathcal{N}}(\cdot)$ are the cumulative  and probability density functions of a standard Gaussian, and $\beta_t \in \Real_+$ must be appropriately chosen.

% \begin{wrapfigure}[10]{R}{.5\textwidth}
%     \vspace{-.5\baselineskip}
% \begin{minipage}{.5\textwidth}
{
\setlength{\textfloatsep}{0cm}
\begin{algorithm}[t]
\setstretch{1.1}
\begin{algorithmic}[1]
\REQUIRE{Set of candidates $\armset$, acq.~func.~$u_0$}
\STATE Initialize $t = 0$
\FOR{$t=\{1, \dots, T\}$}
	\STATE Select $\bx_{t+1} = \argmax_{\bx \in \armset}u_{t}(\bx)$\;
    \STATE Get feedback $y_{t+1}$
    \STATE Update $\mu_t$, $\sigma_t$, $\beta_t$ and $u_t$
\ENDFOR
\end{algorithmic}
\caption{Optimistic GP optimization\label{alg:opt-gp}}
\end{algorithm}
}
% \end{minipage}
% \end{wrapfigure}

Given an acquisition function $u_t$, a standard way to obtain a \gpopt algorithm
based on it is to apply the meta-algorithm reported in \Cref{alg:opt-gp}.
In particular, the acquisition function is used by the meta-algorithm
to choose candidates optimistically/greedily w.r.t.~$u_t$.
Applying \Cref{alg:opt-gp} to the \gpucb and \gpei acquisition function
we obtain exactly the \gpucb and \gpei algorithm,
which are guaranteed to 
achieve low regret \cite{srinivas2010gaussian,wang2014theoretical,chowdhury2017kernelized}.
Unfortunately, optimizing $u_t$ or even evaluating it is computationally inefficient, as
it involves evaluating $\sigma_t(\bx)$ which requires at least $\bigotime(t^2)$ time due to the
multiplication $(\bK_t + \lambda\bI)^{-1}\kerfunc(\bX_t, \bx_j)$.
Moreover, the sequential nature of the optimization protocol precludes the possibility
of running experiments in parallel, which is also necessary to achieve true scalability.
We propose now a new approach to address both bottlenecks.

 \section{Efficient GP Optimization With Few Unique Candidates}\label{sec:bbkb}

Our approach to achieve scalability will be composed by two main ingredients.
The first ingredient is a reformulation of the GP posterior that can be efficiently
computed when the GP is supported on a small number $\uq_t$ of unique candidates (\ie
the history $\bX_t$ contains $t$ rows with repetitions, of which only
$\uq_t$ are unique). To leverage this first ingredient we then need
a second ingredient, a generic approach
to guarantee that $\uq_t$ stays small. This takes the form
of a modification of the usual \gpopt loop of \Cref{alg:opt-gp}
to reduce the number of candidate switches, which acts as an upper bound on
$\uq_t$. Since our switching
rule is agnostic to $u_t$, our approach remains a flexible meta-algorithm
that can be applied to different acquisition functions to obtain different
efficient \gpopt algorithms.
As an added bonus, we also show how these choices allow to seamlessly incorporate
candidate batching in the resulting algorithm, which improves experimental scalability.

{
\setlength{\textfloatsep}{0cm}
\begin{algorithm}[t]
\setstretch{1.1}
\begin{algorithmic}[1]
\REQUIRE{Set of candidates $\armset$, acq.~func.~$u_0$, threshold $C$}
\STATE Initialize $t = 0$
\FOR{$\epoch=\{1, \dots\}$}
	\STATE Select $\bx_{t+1} = \argmax_{\bx \in \armset}u_{t}(\bx)$\;
    \STATE Evaluate $\bx_{t+1}$ for $B_{\epoch} = \lfloor (C^2 - 1)/\sigma_{t}^2(\bx_{t+1}) \rfloor$ times (\ie $\bx_{s} = \bx_{t+1}$ for all $s \in [t+1, t+B_{\epoch}]$)
    \STATE Get feedback $\{y_s\}_{s=t+1}^{t+B_{\epoch}}$
    \STATE Set $t = t + B_{\epoch}$, update posteriors $\mu_t$, $\sigma_t$ and acquisition function $u_t$
\ENDFOR
\end{algorithmic}
\caption{Meta-algorithm to minimize unique candidates (\mumeta)\label{alg:meta-mini}
}
\end{algorithm}
}

\vspace{-0.1in}
\subsection{A meta-algorithm to select few unique candidates}
\vspace{-0.05in}
%In general GP optimization algorithms can select a large number of
%diverse candidates during exploration, and it is very hard to reason and provide
%guarantees on their uniqueness.
%Instead, we can simply upper bound the number of unique candidates with the
%number of candidate switches (\ie number of times $\bx_t \neq \bx_{t+1}$)
%which is easier to algorithmically enforce.
To control the number of candidate switches
we generalize and improve the rarely switching OFUL (\rsoful) algorithm \cite{abbasi2011improved}.
In particular, we extend \rsoful from euclidean to Hilbert spaces, modify it
to operate with acquisition functions different than the \gpucb one,
and replace \rsoful's switching rule with an improved criteria recently
proposed for batched GP optimization \cite{calandriello2020near}.
Our final meta-algorithm \mumeta is reported in \Cref{alg:meta-mini}.

We refer to it as a meta-algorithm
because it can generate different \gpopt algorithms
using different
acquisition functions $u_0(\cdot)$ as input. As an example,
if we use $u_{t}^{\gpucb}(\cdot)$ (\Cref{eq:acq-gpucb}) we obtain an algorithm
similar to \gpucb that we call minimize-unique-\gpucb or \mugpucb for short.
Similarly, we can use \Cref{alg:meta-mini} to convert \gpei into \mugpei.
Moreover, all of the
algorithms generated by \mumeta can compute posteriors efficiently
using a reformulation of $\mu_t$ and~$\sigma_t$ introduced later in
\Cref{eq:mu-post-gp}
rather than the standard formulation of \Cref{eq:post-gp}.

The meta-algorithm operates in epochs/batches indexed by $h$, where each epoch is delimited by a
candidate switch.
%Therefore, ignoring the possibility that the same candidate
%is selected in two different epochs, the current number of epochs $h$ is
%an upper bound $\uq_t \leq h$ on the number of unique candidates.
Inside each epoch, the meta-algorithm selects the next candidate $\bx_{t+1}$ to
be evaluated by maximizing the acquisition function $u_t$, and we discuss more
in detail later how easy or hard this inner optimization problem can be based
on properties of $u_t$ and $\armset$.
Then, after selecting the epoch's candidate
the meta-algorithm must choose the length of the epoch.
Since we do not switch candidate until the end of the epoch, this amounts to selecting
how many times the candidate $\bx_{t+1}$ will be repeatedly evaluated. The main trade-off here is between long epochs that
improve scalability and short epochs that make it easier to switch often and explore/exploit efficiently.
Specializing an approach used in the \bbkb algorithm \cite{calandriello2020near} to our setting, we propose to select $B_{\epoch} = \lfloor ({C^2} - 1)/\sigma_{t}(\bx_{t+1}) \rfloor$,
which as we prove achieves both goals.
Finally, at the end of the epoch we collect the feedback, and use it to update
the GP posterior and the chosen acquisition function. The loop is then repeated
forever or until a desired number of epochs/steps is reached.
%In particular, we specialize a result introduced by
%\citet{calandriello2020near} for their \bbkb algorithm to take advantage that our meta-algorithm sticks
%to the same candidate for the whole epoch, which greatly simplifies
%the termination rule and reduce the cost to compute it.

We also highlight that not all epoch-based \gpopt algorithms
are also batched \gpopt algorithms. In particular, in batched GP algorithms
candidates in a batch can be evaluated in parallel before feedback of previous candidates is available.
Since there is no dependency on the feedback
neither in our candidate selection (it is always the same candidate)
nor in the epoch termination rule, our meta-algorithm
can transform a sequential GP optimization algorithm to a batch variant,
\eg \mugpei is a batch algorithm even though \gpei is not.

\subsection{GPs supported on few unique candidates.}

We can now show that if $h$ is small, all the relevant quantities from \Cref{alg:meta-mini} can be computed efficiently.
In particular, after running \Cref{alg:meta-mini} for $h$ epochs let $\bX_{\epoch}$ indicate the $\epoch \times d$ matrix containing
the $h$ candidates selected so far.
To simplify exposition, we will assume all $h$ candidates are distinct,
and comment in the appendix how removing this assumption just involves
slightly harder book-keeping to merge
feedbacks coming from two identical candidates selected in different epochs.
%Note that the number of unique candidates
%selected at epoch $h$ could be smaller than $h$, \ie two epochs might select the same
%candidate, especially in the later stages of the optimization when the algorithm
%is very confident about which candidate is optimal and should be selected often.
%However, we will ignore this possibility and assume that
%all $h$ candidates in each epoch are distinct.
%This is without loss of generality, as the only difference both in the algorithm
%and the analysis would be a slightly harder book-keeping to merge
%feedbacks coming from two identical candidates selected in different epochs.

Given $\bX_h$, let $\bW_{\epoch} \in \Real^{\epoch \times \epoch}$ be a diagonal matrix;
where $[\bW_{\epoch}]_{i,i} = B_i$ contains the number of times the candidate of the $i$-th epoch
is contained in $\bX_t$ (\ie number of times it is selected),
and let $t_i$ denote the starting time-step of the $i$-th epoch.
Then, we have
\begin{align}
    \begin{aligned}
&{\mu}_{t_h}(\bx) = \kerfunc(\bx, \bX_{\epoch})(\bK_{\epoch} + \lambda\bW_{\epoch}^{-1})^{-1}\by_{\epoch},\\
&
{\sigma}_{t_h}^2(\bx) = \kerfunc(\bx, \bx) - \kerfunc(\bx, \bX_{\epoch})(\bK_{\epoch} + \lambda\bW_{\epoch}^{-1})^{-1}\kerfunc(\bX_{\epoch}, \bx)
\end{aligned}\label{eq:mu-post-gp}
\end{align}
where $\bK_{\epoch} = \kerfunc(\bX_{\epoch}, \bX_{\epoch}) \in \Real^{\epoch \times \epoch}$,
and $\by_{\epoch} \in \Real^{\epoch}$ is such that $[\by_{\epoch}]_i = \sum_{s=t_i}^{t_i + B_i} y_s$,
Using the feature-space view of GPs, it is straightforward to prove using
basic linear algebra identities that the posterior
\Cref{eq:mu-post-gp} is equal
to \Cref{eq:post-gp}, \ie that these are not
approximations to the posterior but \emph{reformulations} of the posterior (see \Cref{sec:app-theory}).
However, when $h \ll t$, Eq.~\ref{eq:mu-post-gp}
can be computed efficiently in $\bigotime(h^3)$, since it only involves
the inversion of an $h \times h$ matrix $\bK_h$.
%We will only consider the reformulated posterior
%at the beginning step $t_h$ of each epoch $h$, since that is all \mumeta requires and simplifies the notation. However
%the same efficient reformulation can be done for any step $t \in [t_i, t_i + B_i]$
%at the only cost of a more complex notation to make the book-keeping consistent
%for the most recent epoch/entry in $\bW_h$ and $\by_h$.

The reformulations of \Cref{eq:mu-post-gp}
are not completely new, e.g., \textcite{picheny2013quantile}
presented similar reformulations based on averaged feedbacks.
At a higher level, they can be seen as a special
case of the more generic framework of reformulating GPs posteriors
as distribution conditioned on a set of inducing points, also
known as sparse GP approximation \cite{quinonero-candela_approximation_2007}.
Traditionally, this framework has been designed to identify a small
number of inducing variables that could act as a bottleneck for the GP
model, reducing its number of parameters and accuracy, but also the
computational cost of inference. However, unlike previous approaches we propose
to utilize the \emph{whole optimization history} as inducing variables,
removing this bottleneck.
Therefore we are not approximating a full GP with a factorized GP, or a dense
GP with a sparse GP, but simply exploiting the intrinsic sparseness
and low number of implicit parameters that arise when GPs
are defined on few unique candidates.
Note that enforcing a candidate selection strategy that keeps $\uq_t$ under control is crucial
in creating a gap between \Cref{eq:post-gp} and \Cref{eq:mu-post-gp}.
This explains why historically this reformulation has not seen
success in fields such as the standard supervised GP regression setting,
where the selection process is not under the control of the learner.
In particular, when $\bX_t$ is sampled i.i.d.~from some
distribution the rows are all different w.p.~1, and our method offers no
scalability improvement. Looking instead at the history of GP optimization
methods, this reformulation has been largely ignored because most approaches
try to evaluate a very diverse sets of candidates to improve initial
exploration. This increases the number of unique historical
candidates, and makes the impact of the reformulation negligible or
even detrimental since it prevents other kinds of efficient incremental
updates.
%The switching condition is crucial in guaranteeing both that
%enough unique candidates are explored to guarantee low regret, but at the same
%time few enough to guarantee high scalability.

 \vspace{-0.1in}
\section{Regret and computational meta-analysis}\label{sec:comparison}
\vspace{-0.1in}

Rather than designing a specialized analysis for every \textsc{mini}-variant,
we propose instead a meta-analysis that cover a generic $u_t$ acquisition
function, and then will instantiate this analysis at the end
of the section with $u_t^{\gpucb}$ and $u_t^{\gpei}$.

While the computational meta-analysis holds for any
$u_t$, the regret meta-analysis requires $u_t$ to satisfy a simple condition.
Given feedback up to step $t-1$, candidates selected
optimistically using the acquisition function $u_t$ at step $t$ must satisfy
w.h.p.~this bound on the instantaneous regret $r_t$
\begin{align}\label{eq:instant-regret}
r_t \triangleq f^{\star} - f(\bx_t) \leq G_t\sigma_{t-1}(\bx_t),
\end{align}
for some non-decreasing sequence $G_t$.
This condition is satisfied for example by \gpucb
\cite{srinivas2010gaussian,abbasi2011improved,chowdhury2017kernelized} and
\gpei \cite{wang2014theoretical}.
Based on this condition, our paper's main result is the following meta-theorem.
\begin{theorem}
    \label{thm:main-meta}
For any $1 < C$, set $\armset$ and acquisition function $u_0$,
\Cref{alg:meta-mini} runs in
$\abigotime(T + \gamma_T\cdot(A + \gamma_T^3))$ time,
$\abigotime(T\gamma_T)$ space,
and $\abigotime(\gamma_T)$ epochs/batches,
% \begin{align*}
%     &\abigotime(T + \gamma_T\cdot(A + \gamma_T^3)) \;\text{ time,}\\
%     &\abigotime(T\gamma_T) \;\text{ space},\\
%     &\abigotime(\gamma_T) \;\text{ epochs/batches},
% \end{align*}
where $A$ indicates the complexity of
solving the inner optimization problem
(i.e., $\argmax_{\bx \in \armset}u_{t}(\bx)$).

Moreover, if $u_t$ satisfies w.p.~$1-\delta$ the condition
of \Cref{eq:instant-regret},
then for all steps $t$ the instantaneous regret $r_t^{\text{\mumeta}}$ of the candidates selected by \Cref{alg:opt-gp}
%\begin{align*}
satisfies $r_t^{\text{\mumeta}} \leq C G_t \sigma_{t-1}(\bx_t)$
for the same sequence of $G_t$ and with the same probability.
%\end{align*}
\end{theorem}
Focusing on regret, \Cref{thm:main-meta} shows that restricting the number
of candidate switches does not greatly impact regret, just as in
\cite{abbasi2011improved}. Moreover \mumeta's improved
switching rule reduces the regret amplification factor from
$\bigotime(2^C)$ of \rsoful to $\bigotime(C)$.
Additionally, our result also highlights for the first time
that approximating the GP model is not necessary to achieve
scalability and no-regret, 
and simply using a different candidate selection strategy
brings great computational savings.
Moreover, compared to existing methods based on efficient reformulations of GPs \cite{picheny2013quantile, binois2019replication} we explicitly target minimizing
unique candidates as a goal of our algorithm, and can
rigorously quantify the computational improvement
that we are guaranteed to achieve compared to a vanilla
GP inference approach.

%Moreover,
%the trade-off between computational cost and regret becomes much
%easier to analyze when focusing on switches.
%Quantifying how restricting $\uq_t$ might affect the performance
%of the learner compared to the general $\uq_t = t$ is a hard combinatorial
%problem. Focusing on switches greatly simplifies the regret guarantees,
%since we only have to consider how
%sub-optimal it is to stick to a candidate for prolonged periods of time,
%when compared to switching at every step.

On the computational side, we bound the number of switches and therefore
of unique candidates using the information gain $\abigotime(\gamma_T)$.
This, allows us to bound both the number of batches and the cost
of evaluating the GP posterior, but unfortunately without knowing
how expensive it is to optimize the acquisition function it is not
possible to fully characterize the runtime. We postpone this to a later
discussion. Nonetheless, an important aspect that can be already included
in the meta-theorem is that this maximization only needs to be performed
once per batch. Previous approaches tried to promote diverse batches,
and therefore this optimization had to be repeatedly performed with a runtime
of at least $\abigotime(TA)$, while for the first time we achieve a runtime $\abigotime(T + A)$ linear in $T$.
However, this crucially relies on being able to 
repeatedly evaluate the same candidate,
which is not possible in all optimization settings.

\subsection{Regret meta-analysis details}\label{sec:meta-regret}
%Without the constraint that the candidate selected have to achieve low regret
%it is trivial to propose extremely scalable algorithms that however over-explore
%or over-exploit.
%Therefore, any consideration on the computational complexity of \Cref{alg:meta-mini}
%can only be done after guaranteeing that if the original algorithm
%achieved low regret, the resulting $\mu$-variant also guarantees low regret.

Starting from the condition in \Cref{eq:instant-regret}, the meta-analysis is
based on well-established tools. First we have to deal with the fact that
\Cref{alg:opt-gp} only updates
the posterior and receives feedback
at the end of the batches.
In particular, this means that \Cref{eq:instant-regret} only applies
for the first evaluation of the candidate, and not across the batch.
Tools developed in the original \rsoful analysis \cite{abbasi2011improved}
and refined in the
context of batch GP optimization \cite{calandriello2020near} can be used to control this.
% \begin{lemma}[{restate=[name=restated]linstaregbound}]\label{lem:instaregbound}
\begin{lemma}\label{lem:instaregbound}
Let $u_t$ be an acquisition function that satisfies \Cref{eq:instant-regret} at
the beginning of each batch. Then running \Cref{alg:meta-mini} with parameter
$C$ and the same $u_t$ guarantees that 
%\begin{align*}
$r_t \leq C G_t \sigma_{t-1}(\bx_t)$
w.p.~$1-\delta$ 
%\end{align*}
for all steps $t$ (i.e., not only at the beginning of each batch).
\end{lemma}

\vspace{-.75\baselineskip}
\begin{proof}[Sketch of proof]
Let $t$ be the first step of epoch/batch $h$ (i.e., we received
feedback up to $t-1$). Then
% \vspace{-.5\baselineskip}
\begin{align*}
    r_{t'}
&\leq G_{t-1}\sigma_{t-1}(\bx_{t'})
    = G_{t-1}\tfrac{\sigma_{t-1}(\bx_{t'})}{\sigma_{t'-1}(\bx_{t'})}\cdot \sigma_{t'-1}(\bx_{t'})\leq G_{t-1} C \cdot \sigma_{t'-1}(\bx_{t'}),
\vspace{-.5\baselineskip}
\end{align*}
where the first inequality comes from applying \Cref{eq:instant-regret}
to step $t'$ since $\bx_t = \bx_{t'}$ (all candidates
in the batch are the same), and the second inequality comes from the fact
that the length of each batch $B_h$ is designed to
guarantee that for any $t' > t$ inside the batch we can bound the ratio
$\sigma_{t-1}(\bx)/\sigma_{t'-1}(\bx) \leq \bigotime(C)$ (details in the appendix).
Therefore, simply by losing a constant factor $C$ we can recover the
regret $r_{t'}$ as if we applied \Cref{eq:instant-regret} with all the feedback
up to step $t'-1$.
\end{proof}
\vspace{-.5\baselineskip}

%Then the remaining steps of the meta-proof follow general and well-established
%steps. Summing the instant regret across $T$ iterations bounds the cumulative
%regret as $R_T^{\text{\mumeta}} \leq \bigotime(\sum_{t=1}^T C \sigma_{t-1}(\bx_t))$.
%Crucially, as the learner collects more feedback the posterior
%$\sigma_{t}(\cdot)$ shrinks, and the sum of posterior variances can be bounded
%using the information gain
%%either $R_t \leq \abigotime(\beta_t \sqrt{T\deff(\bX_t)})$ or
%$R_T^{\text{\mumeta}} \leq \abigotime(\sqrt{CT\gamma_T})$.

%We can formalize 
%The following lemma shows that this blueprint can be straightforwardly extended to include the delayed switching of \Cref{alg:meta-mini}
%as follows.
%This is easily shown by applying \cite[Lem.~3]{calandriello2020near}
%to our special setting where all candidates in the batch are equivalent.
%In particular, this lemma guarantees that for all candidates $t'$ in the
%batch $[t, t+ B_h]$ the ratio $\sigma_{t}(\bx_{t'})/sigma_{t'}$
%This not only simplifies 
%A straightforward consequence of \Cref{l:instaregbound} is that
%$\mu$-\gpucb and all other $\mu$-variants achieve the same regret
%as their original formulation, up to an extra $C$ factor.

Established the $G_t C \sigma_{t-1}(\bx_t)$ bound, the rest of
each regret proofs slightly diverge depending on the acquisition function,
differing mostly on the specific sequence of $G_t$, and we discuss in
\Cref{sec:meta-apply} where we also discuss how the final regret of each of the
alternatives scales.
For now, we would like to highlight that the $1-\delta$ success probability
of \Cref{lem:instaregbound}, and therefore the probability of low regret, depends only on the randomness
of the candidate selection, \ie the same randomness present in the original variants.
This is in stark contrast with existing results for scalable GP optimization
where to provide guarantees on the regret the GP was approximated using
technique relying on some form of additional randomness (e.g., random projections). In other words, our $\mu$-variants are much more \emph{deterministic} compared to other scalable methods.

\subsection{Computational meta-analysis.}\label{sec:meta-comp}
Similarly to regret, we provide here a meta-analysis for a generic
$u_t$, parametrizing the analysis in terms of a
generic $\Narm$ computational cost of optimizing $u_t$ (i.e., computing
$\argmax_{\bx \in \armset}u_{t}(\bx)$)).
With this notation, the computational complexity of \mumeta is
$\bigotime(T + \numepoch \cdot \Narm + \numepoch \cdot \numepoch^3))$.
The $\bigotime(T)$ term covers the time necessary to store/load
the $T$ feedbacks. The $\numepoch \Narm$ term comes from the
$\numepoch$ optimizations of $u_t$, once for each loop.
Finally the $\bigotime(\numepoch \cdot \numepoch^3)$ term is due to the computation of the length
$B_\epoch$ of each batch, which requires at each loop to
computing the posterior $\sigma_t$
using the $\bigotime(\numepoch^3)$ efficient reformulation.
We can refine this analysis along two axes:
bounding $\numepoch$, and bounding $\Narm$. 

Bounding the number of switches $\numepoch$
is fully part of the meta-analysis,
since our bound on the number of epochs does not directly depend
on the specific acquisition function $u_t$.
\begin{lemma}\label{lem:mu-switches}
% \begin{lemma}[{restate=[name=restated]minicomplexity}]\label{lem:mu-switches}
After $T$ steps, \mumeta performs at most 
$h \leq \bigotime\left(\frac{C^2}{C^2-1}\gamma_T\right)$ switches.
\end{lemma}
Notice that once again this is a deterministic statement, since unlike existing
methods our approach does not depend on randomized methods to increase
scalability. Combining this result in the meta-analysis we obtain
an overall runtime of $\bigotime(T + \gamma_T\cdot(\Narm + \gamma_T^3))$. 

Bounding the optimization cost $\Narm$ can only be done
by making assumptions on $u_t$ and $\armset$, since
in general maximizing the acquisition function is a non-convex
optimization problem, often NP-hard. In
the simpler case where $\armset$ is finite with cardinality $|\armset|$,
and $u_t$ is based only on $\mu_t$ and $\sigma_t$ of each candidate,
then the runtime becomes
$\bigotime(T + \numepoch\cdot(|\armset|\numepoch^2 + \numepoch^3)$.
    Further combining
this with \Cref{lem:mu-switches} we obtain an overall runtime of
$\bigotime(T + |\armset|\gamma_T^3 + \gamma_T^4)$. We can compare this
result with the $\abigotime(T|\armset|\gamma_T^2)$ runtime complexity
of the currently fastest no-regret algorithm, \bbkb \cite{calandriello2020near}
to see that the potential runtime improvement is large (i.e., from
$\bigotime(T|\armset|)$ to $\bigotime(T + |\armset|\numepoch)$.
This is because \mumeta only needs to maximize $u_t$ once per batch,
while existing batched algorithms need multiple maximization per batch
to promote in-batch diversity, using different strategies such as feedback
hallucination \cite{desautels_parallelizing_2014}.
Even if we ignored the cost of optimizing $u_t$, our result would still be the first to decouple the the $\bigotime(T)$ component from $\gamma_T$.
This is because for existing batching rule, even computing the batch length required to evaluate at least one $\sigma_t$ at each step, with each evaluation costing at least $\bigotime(\gamma_T^2)$.
Instead, our batching rule is based only on $\sigma_t$ once at the beginning of the batch.

\subsection{Instantiating the meta-analysis}\label{sec:meta-apply}
Instantiating the results from the previous section we can now provide
bounds for two popular acquisition functions. For simplicity
and to guarantee that all steps of the algorithm can be implementable
in accord with the regret analysis, we restrict ourselves to
the assumption of finite $\armset$ where the acquisition functions can be exactly
maximized.
For all of these variants, the run-time is bounded as
$\bigotime(T + |\armset|\gamma_T^3 + \gamma_T^4)$, so we will mainly discuss
regret in this section.

\paragraph{\mugpucb}
We consider frequentist ($\normsmall{f} \leq \fnorm$)
and Bayesian ($f \sim GP$) settings.
\begin{theorem}\label{thm:mu-gp-ucb-freq}
    Assume $\normsmall{f} \leq \fnorm$. For any $1 < C$, $\delta \in [0, 1]$, run
\Cref{alg:opt-gp} with
$u_t = u_t^{\gpucb}$, $\lambda = \xi^2$, and
\begin{align*}
\displaystyle
& \beta_h = \Theta
\left(\sqrt{\logdet\left(\tfrac{\bW_h^{1/2}\bK_h\bW_h^{1/2}}{\lambda} + \bI\right) + \log(\tfrac{1}{\delta})} + F\right).
\end{align*}
Then w.p.~$1-\delta$,
$R_T \leq \abigotime\left((\sqrt{\gamma_T} + F)C\sqrt{\gamma_T T}\right)$.
\end{theorem}
This result is a combination of \Cref{thm:main-meta} with {Thm.~1} in \cite{chowdhury2017kernelized}.
The proof of this result is straightforward using \Cref{thm:main-meta}.
In particular, the original \gpucb provides a bound on the regret of
the form $R_T \leq \abigotime(\beta_T\sum_{t=1}^T\sigma_{t-1}(\bx_t))$,
which using \cref{lem:instaregbound} easily becomes $R_T \leq \abigotime(\beta_TC\sum_{t=1}^T\sigma_{t-1}(\bx_t))$, only a $C$ factor worse but with a much
lower computational complexity.
Compared to other approximations of \gpucb, \mugpucb achieve logarithmic
improvements due to a tighter $\beta_h$. In particular, all
previous approximate \gpucb variants had to approximate
$\logdet(\bK_t/\lambda + \bI)$ with a more or less loose upper bound, which
resulted in excessive exploration and worse regret. Instead, \mugpucb
uses the exact log-determinant of the GP, since it only needs to be defined
on the unique candidates.

\begin{theorem}\label{thm:mu-gp-ucb-bayes}
    Assume $f \sim GP$. For any $1 < C$, $\delta \in [0, 1]$, run
\Cref{alg:opt-gp} with $u_t = u_t^{\gpucb}$, setting $\lambda = \xi^2$ and $\beta_t = \sqrt{2 \log(|\armset|t^2\pi^2
/(6\delta))}$. Then w.p.~$1-\delta$,
\begin{align*}
R_T \leq \abigotime(C\sqrt{T\gamma_T})).
\end{align*}
\end{theorem}
This result is a combination of \Cref{thm:main-meta} and {Thm.~1} of \cite{srinivas2010gaussian}.
The main advantage of the Bayesian tuning of \mugpucb is that
computing $\beta_t$ does not require to know a bound on the norm of the function $F$ (which is often infinite
under the GP prior). However the algorithm still requires access
to the noise level $\xi$ to tune the $\lambda$ parameter.

%\subsection{$\mu$-\gpts}
%Frequentist bounds can be obtained starting from \citet{chowdhury2017kernelized}.

\paragraph{\mugpei}
We can combine \Cref{thm:main-meta} with {Thm.~1} of \cite{wang2014theoretical} under frequentist assumptions.
\begin{theorem}\label{thm:mu-gp-ei-freq}
    Assume $\normsmall{f} \leq \fnorm$. For any $1 < C$ and $\delta \in [0, 1]$, run
\Cref{alg:opt-gp} with $u_t = u_t^{\gpei}$, $\lambda = \xi^2$, and
$\beta_h = 
\big(\logdet(\bW_h^{1/2}\bK_h\bW_h^{1/2}/\lambda + \bI) + \sqrt{\logdet(\bW_h^{1/2}\bK_h\bW_h^{1/2}/\lambda + \bI)\log(\tfrac{t}{\delta})} +\log(\tfrac{t}{\delta})\big)^{1/2}$.
Then w.p.~$1-\delta$,
$R_T \leq \abigotime\left((\sqrt{\gamma_T} + F)C\sqrt{\gamma_T T}\right)$.
\end{theorem}

Again we can easily integrate \Cref{thm:main-meta} in the original
proof. In particular, the original \gpei analysis provides
a bound on the regret of the form
$R_T \leq \abigotime\left(\left(\sqrt{\lambda}F + \beta_T\right)\sum_{t=1}^T\sigma_{t-1}(\bx_t)\right)$,
which again using \cref{lem:instaregbound} easily becomes $R_T \leq \abigotime\left(\left(\sqrt{\lambda}F + \beta_T\right)C\sum_{t=1}^T\sigma_{t-1}(\bx_t)\right)$ recovering the original regret up to constants. Although the regret
of \mugpucb and \mugpei are comparable, the underlying algorithms have important
differences. In particular, tuning $\beta_h$ in \mugpei does not require knowing
a bound $F$ on the norm of the function, which is hard to obtain in many cases.
However the analysis of \mugpei requires to set $\lambda = \xi^2$, which
might be as hard to estimate.

\section{Experiments}\label{sec:exp}

\newcommand{\vcenteredinclude}[1]{\begingroup
    \setbox0=\hbox{\includegraphics[height=.75\baselineskip]{#1}}%
\parbox{\wd0}{\box0}\endgroup}
\begin{figure*}[t]
    \centering
     \small{\vcenteredinclude{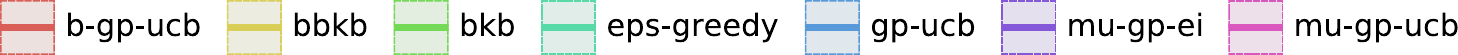}}

     \vspace{.5\baselineskip}
	 \begin{subfigure}[b]{0.32\textwidth}
		 \centering
         \includegraphics[width=\textwidth,trim=.2cm .2cm 4cm .2cm,clip]{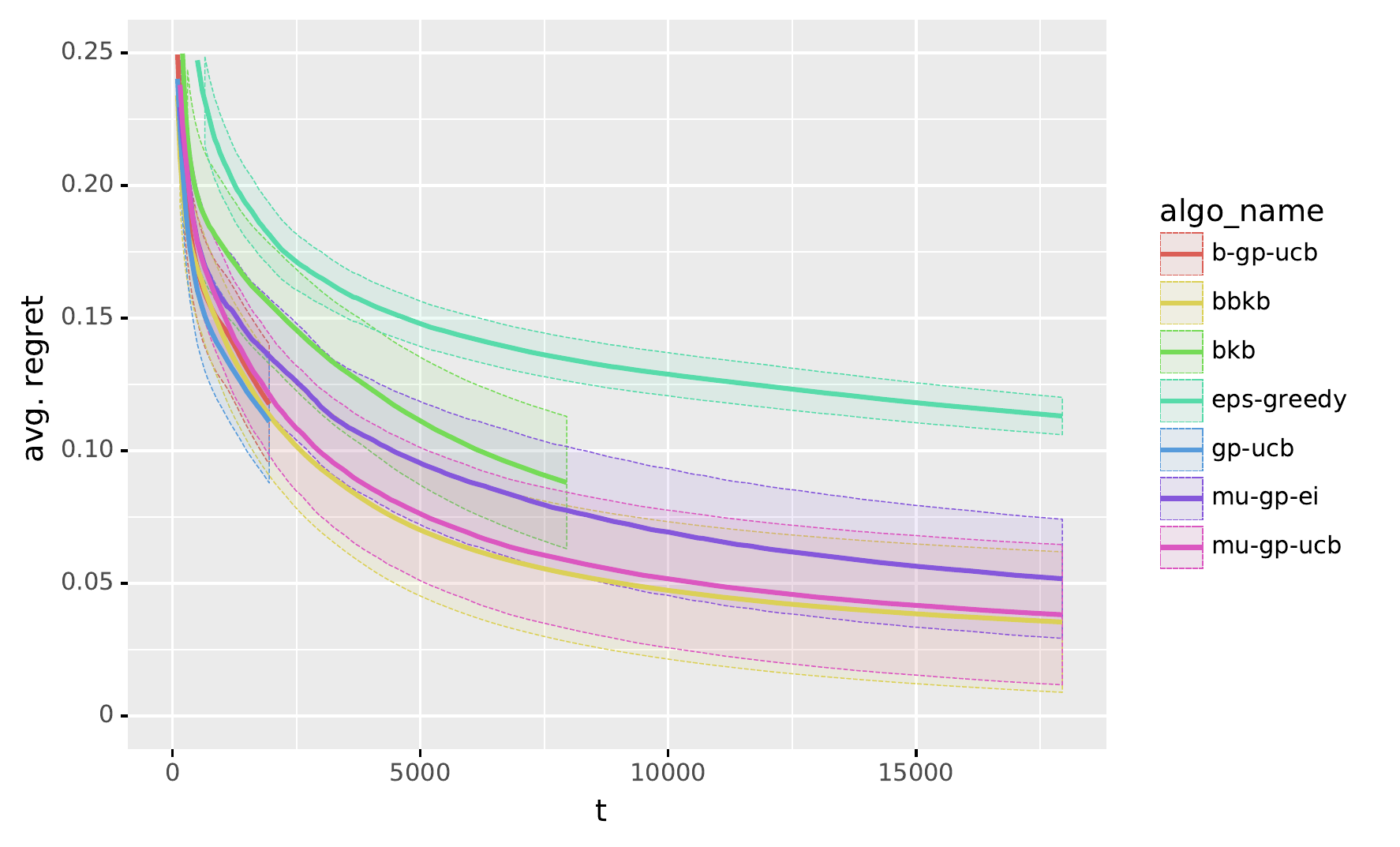}
		 \caption{$R_t/t$ against step $t$}
		 \label{fig:regret}
	 \end{subfigure}
	 \hfill
	 \begin{subfigure}[b]{0.32\textwidth}
		 \centering
	\includegraphics[width=\textwidth,trim=.2cm .2cm 4cm .2cm,clip]{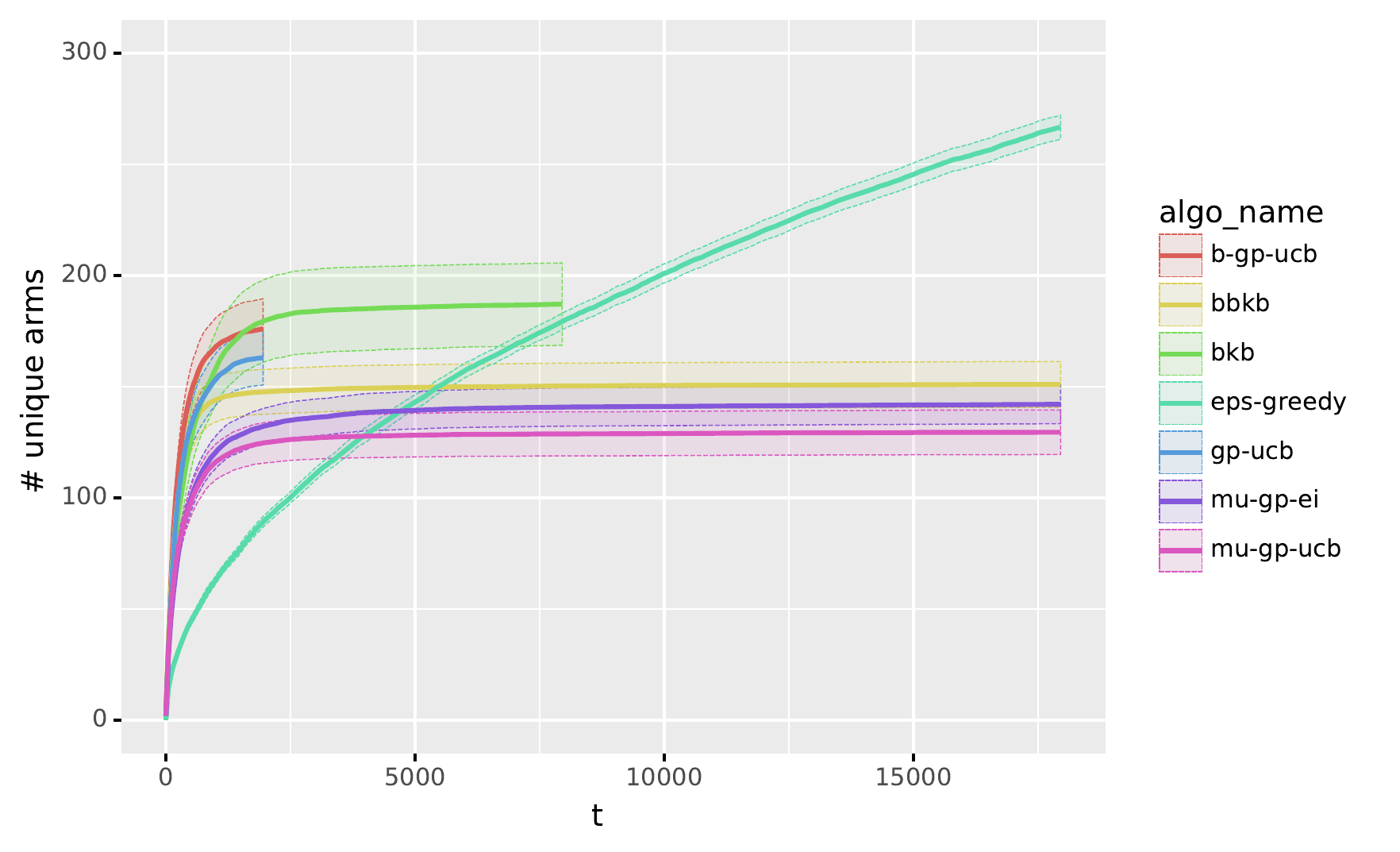}
		 \caption{$\uq_t$ against step $t$ }
		 \label{fig:unique}
	 \end{subfigure}
	 \hfill
	 \begin{subfigure}[b]{0.32\textwidth}
		 \centering
	\includegraphics[width=\textwidth,trim=.2cm .2cm 4cm .2cm,clip]{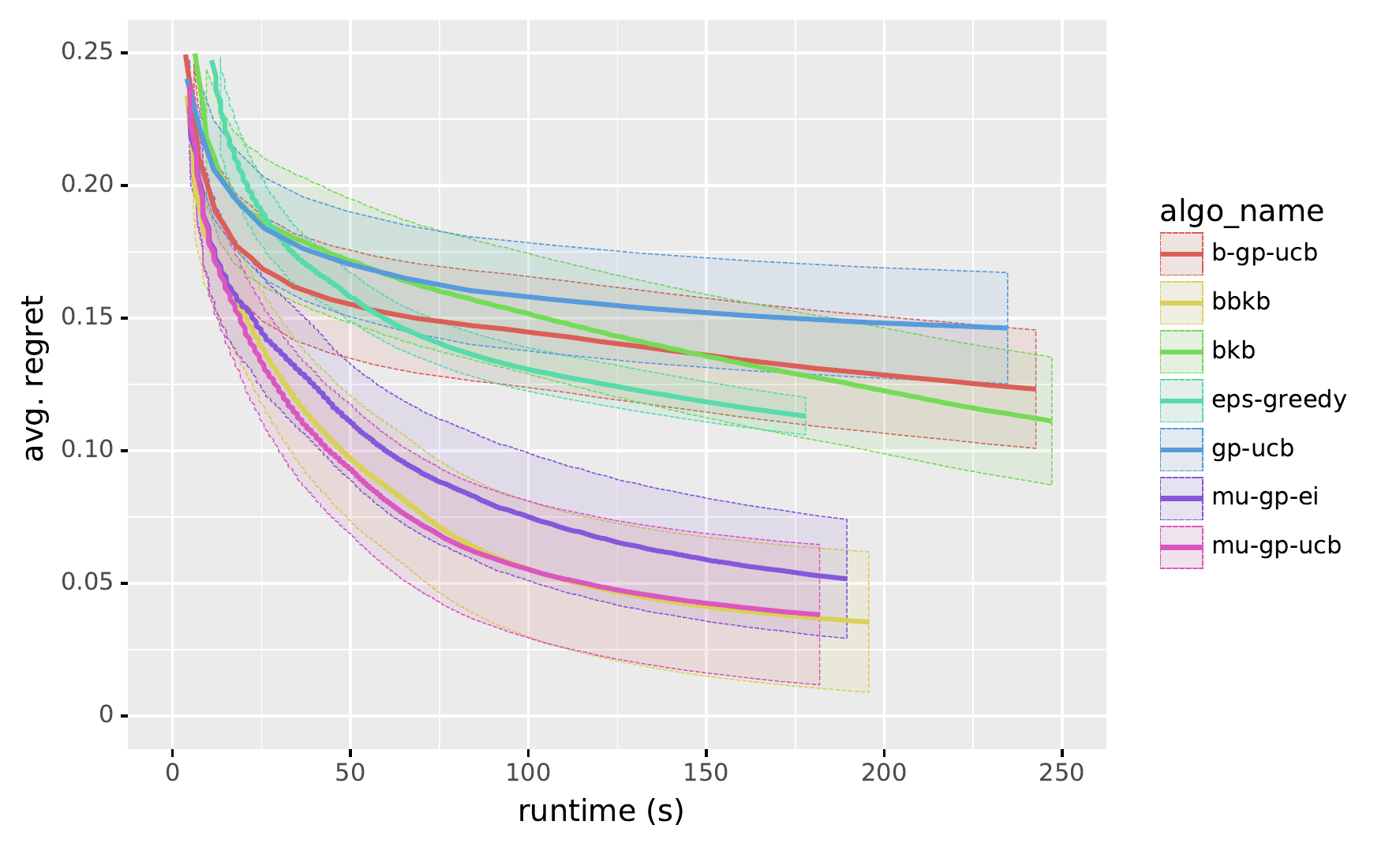}
    \caption{$R_T$ against wall-clock time (s) }
		 \label{fig:wallclock}
	 \end{subfigure}
\end{figure*}
We now evaluate our proposed approach empirically, focusing
here on real-world data and on synthetic data in the appendix.
In particular, following the approach taken in
\cite{calandriello2020near} we evaluate \mugpucb
and \mugpei on NAS-bench \cite{ying2019bench}.
%and on a hyperparameter
%tuning dataset from \citet{metz2020using} that we refer to as task-bench.
We compare \mugpucb and \mugpei with \bbkb \cite{calandriello2020near},
\gpucb \cite{srinivas2010gaussian}, \bgpucb \cite{desautels_parallelizing_2014},
\bgpts \cite{kandasamy2018parallelised}, and an epsilon greedy strategy.
For \mugpucb, \mugpei, and \bbkb hyperparameters that need to be tuned are
$C$, the bandwidth of the Gaussian kernel $\sigma$, while
$\lambda$ is set to the 90-th percentile of the standard deviation
of the target as an oracle for the noise level.
For epsilon greedy epsilon
is set as $a/t^b$ and $a \in \{0.1, 1, 10\}$ and $b \in \{1/3, 1/2, 1, 2\}$
is tuned via grid search.
For each experiment we repeat the run 40 times with
different seeds, and report mean and confidence intervals
for the hyperparameter configuration (searched in a grid)
that achieves lowest average regret (specific values reported in the appendix). All experiments are ran using a single, recent generation CPU core to avoid inconsistencies between some implementations using paralallel BLAS and some not.
%
% \subsection{Nas-bench}
The neural architecture search
setting is the same as \cite{calandriello2020near}. In particular,
the search space is the discrete combinatorial space of possible
1-hidden layer 4 node networks using either convolutions or max-pooling,
that is then used as a module in an inception-like architecture.
The final space $\armset$ has $|\armset| = 12416$ candidates in $d = 19$
dimensions.

In Fig.~(\subref{fig:regret}) we report average regret vs steps (i.e., $t$). To compensate
scale effect in the rewards, the average regret
is normalized by the average regret achieved by a completely exploratory
policy that selects candidates uniformly at random. Both \textsc{mini}-variants
are comparable to the current state of the art \bbkb, and better than
a tuned epsilon greedy.
In Fig.~(\subref{fig:unique}) we report the number of unique candidates (note, not of switches)
selected by each algorithm. We note that despite not explicitly
being designed for this, \bbkb does not select a very large number of
unique arms. However, \mugpucb and \mugpei still select an even
smaller number of unique candidates.
In Fig.~(\subref{fig:wallclock}) we report average regret vs wallclock runtime rather than steps.
Therefore a more efficient algorithm will terminate faster and
achieve a lower average regret in the same span of time.
We see that both \mugpucb and \mugpei are faster than \bbkb,
terminating earlier.
However the empirical runtime gap does not seem to match the large theoretical runtime gap
(i.e., $\bigotime(T + A)$ vs $\bigotime(TA)$). In particular, the actual empirical diversity in \bbkb's
batches seem to be limited, and allows for a large degree of lazy updates, which are not accounted for in the
worst-case complexity.

% \subsection{Task-bench}
% \todod{Luke Metz (main author of task-bench and Googler) is currently re-running
% th task-bench data generation code to have access to a larger candidate space.
% The code itself is open-sourced on
% \url{https://github.com/google-research/google-research/tree/master/task_set}
% and he used it to generate more learning curves that he plans to release after
% this. As soon as we have the data we will proceed with generating new plots.

% The reason the tuned greedy seems to be performing extremely well in this
% case is because due to a few outliers, normalizing the rewards in $[0,1]$
% makes the gap between the average arm and the optimal arm extremely small.
% It will probably be necessary to exclude or otherwise remove these
% outliers to remove this artifact.}
% \includegraphics[width=.32\textwidth]{taskbench_step_vs_regret.pdf}
%     \includegraphics[width=.32\textwidth]{taskbench_step_vs_unique.pdf}
%     \includegraphics[width=.32\textwidth]{taskbench_time_vs_regret.pdf}
\section{Conclusions, limitations and open questions}
Our paper highlighted how existing tools can be combined
in a new effective approach, capable of positively impacting the GP optimization
setting both theoretically and empirically. Theoretically,
as to the best of our knowledge, our proposed \textsc{mini}-variants
achieve the tightest guarantees both in regret and runtime among
scalable GP optimization methods, using a very different approach
that does not require approximating the GP or other randomized approximations.
Empirically, because our method comes with a number of practical properties,
including scalability, adaptability to different acquisition functions,
and being suitable to batched settings
as well as settings where
switching costs are be significant.
However, there remain several limitations, which brings with them open questions.

Our \textsc{mini}-variants inherit the limitations of the
original methods and while some (e.g., scalability) are removed, others remain. In
particular, \mugpucb and \mugpei still require knowledge of quantities that are hard to estimate,
 such as the function norm, noise level, kernel choice, or kernel bandwidth. It is unclear how an on-line
tuning of these quantities might be done without losing the no-regret
guarantees. It would be interesting to see if
recent
approaches to tune these quantities for traditional GP optimization \cite{Berkenkamp2019,durand2018streaming}
can be transferred to \textsc{mini}-variants, leveraging their
unique property of being based on exact GP inference.
Furthermore, while \mumeta could be applied to other acquisition functions, not all would result in a scalable
algorithm. For example \gpts \cite{chowdhury2017kernelized} also satisfies
\Cref{eq:instant-regret}. However, sampling a TS
posterior is not scalable (i.e., $\bigotime(|\armset|^2)$
for finite $\armset$) even for sparse posteriors,
and therefore a hypotetical \textsc{mini}-\gpts would remain not
scalable.

From a complexity perspective,  optimizing the acquisition function exactly
remains one of the hardest obstacles in GP optimization,
and it is still not clear how the no-regret guarantees can be extended
to approximate maximization.
Generic
non-convex optimizers such as DiRECT \cite{mutny_efficient_2018,Jones2001} have an exponential
complexity,
and even considering the effective dimensionality reduction
induced by \Cref{eq:mu-post-gp}, which reduces
the parameter space from $\bigotime(t)$ to $\bigotime(\uq_t)$,
they might remain infeasible.

Despite not being explicitly optimized for this
task, we empirically observe that other approximate \gpopt methods, such as \bbkb, also tend
to select a small number of unique candidates. Indeed, any algorithm that quickly converges to a small set
of good candidates would evaluate a small number of unique candidates.
Therefore, it is an important open question
to try and bound the number of candidates evaluated by a generic
GP optimization algorithm without low-switching enforcement, as it might
show that the reformulations of \Cref{eq:mu-post-gp} might be more broadly applicable than expected.

Finally, looking at the applicability of our method, our whole approach is based on
the possibility of evaluating multiple times the same candidate without
affecting the outcome. While this is a strength in settings with high
switching costs, it also makes
it unsuitable in settings where this is impossible (e.g., repeated medical
testing of a single patient), or limited (e.g., news change over time and cannot be recommended over and over). More subtly, it also makes it poorly suited
to low or noiseless settings, where multiple evaluations of the same candidate
would not be very useful.
% Finally in simple optimization problems that can be
% solved in a few iterations the problem of scalability does not present
% itself and our meta-approach would represent just added complexity on top
% of the original approaches.

\newpage

\section*{Acknowledgements}
This material is based upon work supported by the Center for Brains, Minds and Machines (CBMM), funded by NSF STC award CCF-1231216, and the Italian Institute of Technology. L. R. acknowledges the financial support of the European Research Council (grant SLING 819789), the AFOSR projects FA9550-18-1-7009, FA9550-17-1-0390 and BAA-AFRL-AFOSR-2016-0007 (European Office of Aerospace Research and Development), and the EU H2020-MSCA-RISE project NoMADS - DLV-777826.

\printbibliography

@InProceedings{pmlr-v130-vakili21a,
  title = 	 { On Information Gain and Regret Bounds in Gaussian Process Bandits },
  author =       {Vakili, Sattar and Khezeli, Kia and Picheny, Victor},
  booktitle = 	 {Proceedings of The 24th International Conference on Artificial Intelligence and Statistics},
  pages = 	 {82--90},
  year = 	 {2021},
  editor = 	 {Banerjee, Arindam and Fukumizu, Kenji},
  volume = 	 {130},
  series = 	 {Proceedings of Machine Learning Research},
  month = 	 {13--15 Apr},
  publisher =    {PMLR},
}

@Inbook{Jones2001,
author="Jones, Donald R.",
editor="Floudas, Christodoulos A.
and Pardalos, Panos M.",
title="Direct global optimization algorithmDirect Global Optimization Algorithm",
bookTitle="Encyclopedia of Optimization",
year="2001",
publisher="Springer US",
address="Boston, MA",
pages="431--440",
isbn="978-0-306-48332-5",
doi="10.1007/0-306-48332-7_93",
url="https://doi.org/10.1007/0-306-48332-7_93"
}

@article{durand2018streaming,
  title={Streaming kernel regression with provably adaptive mean, variance, and regularization},
  author={Durand, Audrey and Maillard, Odalric-Ambrym and Pineau, Joelle},
  journal={The Journal of Machine Learning Research},
  volume={19},
  number={1},
  pages={650--683},
  year={2018},
  publisher={JMLR. org}
}

@article{Berkenkamp2019,
  author  = {Felix Berkenkamp and Angela P. Schoellig and Andreas Krause},
  title   = {No-Regret Bayesian Optimization with Unknown Hyperparameters},
  journal = {Journal of Machine Learning Research},
  year    = {2019},
  volume  = {20},
  number  = {50},
  pages   = {1-24},
  url     = {http://jmlr.org/papers/v20/18-213.html}
}

@article{gardner2018gpytorch,
  title={Gpytorch: Blackbox matrix-matrix Gaussian process inference with GPU acceleration},
  author={Gardner, Jacob R and Pleiss, Geoff and Bindel, David and Weinberger, Kilian Q and Wilson, Andrew Gordon},
  journal={Advances in Neural Information Processing Systems},
  volume={2018},
  pages={7576--7586},
  year={2018}
}

@article{wang2014theoretical,
  title={Theoretical analysis of Bayesian optimisation with unknown Gaussian process hyper-parameters},
  author={Wang, Ziyu and de Freitas, Nando},
  journal={arXiv preprint arXiv:1406.7758},
  year={2014}
}

@article{mockus1989bayesian,
  title={Bayesian approach to global optimization: theory and applications},
  author={Mockus, Jonas},
  year={1989},
  publisher={Springer}
}

@inproceedings{calandriello2020near,
  title={Near-linear Time Gaussian Process Optimization with Adaptive Batching and Resparsification},
  author={Calandriello, Daniele and Carratino, Luigi and Valko, Michal and Lazaric, Alessandro and Rosasco, Lorenzo},
  booktitle={International Conference on Machine Learning},
  year={2020}
}

@book{lattimore2020bandit,
  title={Bandit algorithms},
  author={Lattimore, Tor and Szepesv{\'a}ri, Csaba},
  year={2020},
  publisher={Cambridge University Press}
}

@inproceedings{kandasamy2018parallelised,
  title={Parallelised bayesian optimisation via thompson sampling},
  author={Kandasamy, Kirthevasan and Krishnamurthy, Akshay and Schneider, Jeff and P{\'o}czos, Barnab{\'a}s},
  booktitle={International Conference on Artificial Intelligence and Statistics},
  pages={133--142},
  year={2018}
}

@inproceedings{calandriello_2019_coltgpucb,
	Author = {Calandriello, Daniele and Carratino, Luigi and Lazaric, Alessandro and Valko, Michal and Rosasco, Lorenzo},
	Booktitle = {Conference on Learning Theory},
	Title = {Gaussian Process Optimization with Adaptive Sketching: Scalable and No Regret},
	Year = {2019}}

@article{desautels_parallelizing_2014,
	title = {Parallelizing exploration-exploitation tradeoffs in {Gaussian} process bandit optimization},
	volume = {15},
	number = {1},
	journal = {The Journal of Machine Learning Research},
	author = {Desautels, Thomas and Krause, Andreas and Burdick, Joel W.},
	year = {2014},
	pages = {3873--3923},
}

@inproceedings{kathuria_batched_2016,
	title = {Batched gaussian process bandit optimization via determinantal point processes},
	booktitle = {Advances in {Neural} {Information} {Processing} {Systems}},
	author = {Kathuria, Tarun and Deshpande, Amit and Kohli, Pushmeet},
	year = {2016},
	pages = {4206--4214},
}

@article{ying2019bench,
  title={Nas-bench-101: Towards reproducible neural architecture search},
  author={Ying, Chris and Klein, Aaron and Real, Esteban and Christiansen, Eric and Murphy, Kevin and Hutter, Frank},
  journal={arXiv preprint arXiv:1902.09635},
  year={2019}
}

@article{li2010contextual,
author = {Li, Lihong and Chu, Wei and Langford, John and Schapire, Robert E.},
institution = {ACM},
journal = {International World Wide Web Conference},
publisher = {ACM Press},
title = {{A contextual-bandit approach to personalized news article recommendation}},
url = {http://rob.schapire.net/papers/www10.pdf},
year = {2010}
}

@inproceedings{scarlett2017lower,
  title={Lower Bounds on Regret for Noisy Gaussian Process Bandit Optimization},
  author={Scarlett, Jonathan and Bogunovic, Ilija and Cevher, Volkan},
  booktitle={Conference on Learning Theory},
  pages={1723--1742},
  year={2017}
}

@article{quinonero-candela_approximation_2007,
	title = {Approximation methods for gaussian process regression},
	urldate = {2017-05-10},
	journal = {Large-scale kernel machines},
	author = {Quinonero-Candela, Joaquin and Rasmussen, Carl Edward and Williams, Christopher KI},
	year = {2007},
	pages = {203--224},
}

@InProceedings{daxberger17a,
  title = 	 {Distributed Batch {G}aussian Process Optimization},
  author = 	 {Erik A. Daxberger and Bryan Kian Hsiang Low},
  booktitle = 	 {Proceedings of the 34th International Conference on Machine Learning},
  pages = 	 {951--960},
  year = 	 {2017},
  editor = 	 {Doina Precup and Yee Whye Teh},
  volume = 	 {70},
  series = 	 {Proceedings of Machine Learning Research},
  address = 	 {International Convention Centre, Sydney, Australia},
  month = 	 {06--11 Aug},
  publisher = 	 {PMLR},
}

@incollection{mutny_efficient_2018,
	title = {Efficient {High} {Dimensional} {Bayesian} {Optimization} with {Additivity} and {Quadrature} {Fourier} {Features}},
	booktitle = {Advances in {Neural} {Information} {Processing} {Systems} 31},
	publisher = {Curran Associates, Inc.},
	author = {Mutny, Mojmir and Krause, Andreas},
	editor = {Bengio, S. and Wallach, H. and Larochelle, H. and Grauman, K. and Cesa-Bianchi, N. and Garnett, R.},
	year = {2018},
	pages = {9019--9030},
}

@article{picheny2013quantile,
  title={Quantile-based optimization of noisy computer experiments with tunable precision},
  author={Picheny, Victor and Ginsbourger, David and Richet, Yann and Caplin, Gregory},
  journal={Technometrics},
  volume={55},
  number={1},
  pages={2--13},
  year={2013},
  publisher={Taylor \& Francis}
}

@article{binois2019replication,
  title={Replication or exploration? Sequential design for stochastic simulation experiments},
  author={Binois, Micka{\"e}l and Huang, Jiangeng and Gramacy, Robert B and Ludkovski, Mike},
  journal={Technometrics},
  volume={61},
  number={1},
  pages={7--23},
  year={2019},
  publisher={Taylor \& Francis}
}

@article{saito2020open,
  title={Open Bandit Dataset and Pipeline: Towards Realistic and Reproducible Off-Policy Evaluation},
  author={Saito, Yuta and Shunsuke, Aihara and Megumi, Matsutani and Yusuke, Narita},
  journal={NeurIPS2021 Datasets and Benchmarks Track},
  year={2021}
}

@book{rasmussen_gaussian_2006,
	address = {Cambridge, Mass},
	series = {Adaptive computation and machine learning},
	title = {Gaussian processes for machine learning},
	isbn = {978-0-262-18253-9},
	publisher = {MIT Press},
	author = {Rasmussen, Carl Edward and Williams, Christopher K. I.},
	year = {2006},
	note = {OCLC: ocm61285753},
}

@inproceedings{chowdhury2017kernelized,
  title={On Kernelized Multi-armed Bandits},
  author={Chowdhury, Sayak Ray and Gopalan, Aditya},
  booktitle={International Conference on Machine Learning},
  pages={844--853},
  year={2017}
}

@inproceedings{abbasi2011improved,
  title={Improved algorithms for linear stochastic bandits},
  author={Abbasi-Yadkori, Yasin and P{\'a}l, D{\'a}vid and Szepesv{\'a}ri, Csaba},
  booktitle={Advances in Neural Information Processing Systems},
  pages={2312--2320},
  year={2011}
}

@inproceedings{srinivas2010gaussian,
	Author = {Srinivas, Niranjan and Krause, Andreas and Seeger, Matthias and Kakade, Sham M},
	Booktitle = {International Conference on Machine Learning},
	Pages = {1015--1022},
	Title = {Gaussian Process Optimization in the Bandit Setting: No Regret and Experimental Design},
	Year = {2010}}

%%%%%%%%%%%%%%%%%%%%%%%%%%%%%%%%%%%%%%%%%%%%%%%%%%%%%%%%%%%%

%%%%%%%%%%%%%%%%%%%%%%%%%%%%%%%%%%%%%%%%%%%%%%%%%%%%%%%%%%%%

\newpage

\appendix

\section{Proofs of \Cref{sec:bbkb}}\label{sec:app-theory}

For several of the proofs in this section it will be useful
to introduce the so-called feature space formulation of a GP posterior
\cite{rasmussen_gaussian_2006}. In particular, to every kernel function
$\kerfunc$ we can associate a feature map $\varphi(\cdot) : \armset \rightarrow \rkhs$
where $\rkhs$ is the reproducing kernel Hilbert space associated with $\kerfunc$ and the GP.
The main property of $\varphi(\cdot)$ is that for any
$\bx$ and $\bx'$ we have $\kerfunc(\bx, \bx') = \varphi(\bx)^\transp\varphi(\bx')$.
With a slight abuse of notation, we will also indicate with $\varphi(\bX) =
[\varphi(\bx_1), \dots, \varphi(\bx_t)]^\transp$ the linear operator obtained
by stacking together the various $\varphi(\cdot)$, such that
$\bK_t = \varphi(\bX_t)\varphi(\bX_t)^\transp$. Note that this also
allows us to define the equivalent of $\bK_t$ in $\rkhs$ as
$\varphi(\bX_t)^\transp\varphi(\bX_t) : \rkhs \rightarrow \rkhs$.
Finally, to connect $\bK_t$ and $\varphi(\bX_t)^\transp\varphi(\bX_t)$ we will
heavily use this fundamental linear algebra equality
\begin{align}\label{eq:prim-dual-eq}
\bB^\transp(\bB\bB^\transp + \lambda\bI)^{-1}\bB = \bB^\transp\bB(\bB^\transp\bB + \lambda\bI)^{-1}
\end{align}
which can be easily shown to be valid for any linear operator $\bB$ using its singular
value decomposition.

\subsection{Proof of equivalence between \Cref{eq:post-gp} and \Cref{eq:mu-post-gp}}
Assume for now that $t$ is the step at the end (\ie $\fb{t} = t$) of batch $h$.
We will relax this assumption at the end of the section to discuss
how this can be extended to intermediate steps.
Using the feature-space representation of a GP (see e.g., {Eq.~(2.11)}in \cite{rasmussen_gaussian_2006}),
and defining
$\bV_t \triangleq \varphi(\bX_t)^\transp\varphi(\bX_t) + \lambda\bI$,
we can rewrite the posterior
variance as
\begin{align*}
{\sigma}_t^2(\bx_i) &= \kerfunc(\bx_i, \bx_i) - \kerfunc(\bx_i, \bX_t)(\bK_t + \lambda\bI)^{-1}\kerfunc(\bX_t, \bx_i)\\
&\stackrel{a}{=} \varphi(\bx_i)^\transp\varphi(\bx_i) - \varphi(\bx_i)^\transp\varphi(\bX_t)^\transp(\varphi(\bX_t)\varphi(\bX_t)^\transp + \lambda\bI)^{-1}\varphi(\bX_t)\varphi(\bx_i)\\
&\stackrel{b}{=} \varphi(\bx_i)^\transp\varphi(\bx_i) - \varphi(\bx_i)^\transp\varphi(\bX_t)^\transp\varphi(\bX_t)(\varphi(\bX_t)^\transp\varphi(\bX_t) + \lambda\bI)^{-1}\varphi(\bx_i)\\
&\stackrel{c}{=} \varphi(\bx_i)^\transp\varphi(\bx_i) - \varphi(\bx_i)^\transp(\underbracket{\varphi(\bX_t)^\transp\varphi(\bX_t) + \lambda\bI}_{\bV_t} - \lambda\bI)\underbracket{(\varphi(\bX_t)^\transp\varphi(\bX_t) + \lambda\bI)^{-1}}_{\bV_{t}^{-1}}\varphi(\bx_i)\\
&\stackrel{d}{=} \varphi(\bx_i)^\transp\varphi(\bx_i) - \varphi(\bx_i)^\transp(\bI - \lambda\bV_{t}^{-1})\varphi(\bx_i) = \lambda\varphi(\bx_i)^\transp\bV_{t}^{-1}\varphi(\bx_i)
= \varphi(\bx_i)^\transp\bA_{t}^{-1}\varphi(\bx_i),
\end{align*}
where in each passage
\begin{itemize}
    \item[$a$)] we simply apply the definition of $\kerfunc$ and $\varphi$;
\item[$b$)] we apply \Cref{eq:prim-dual-eq} with $\varphi(\bX)$ as $\bB$;
\item[$c$)] we add and subtract $\lambda\bI$ to highlight
the presence of $\bV_t$ in the reformulation;
\item[$d$)] we collect $\lambda$ to replace $\bV_t$ with 
$\bA_t \triangleq \varphi(\bX)^\transp\varphi(\bX)/\lambda + \bI$ as {Eq.~(2.11)} in \cite{rasmussen_gaussian_2006}.
\end{itemize}
Exploiting the fact that all candidate in a batch are identical
(i.e., $\bx_{\fb{s} + 1} = \bx_{s}$), and denoting with $\{\bx_j\}_{j=1}^h$
the candidate in each batch we can rewrite $\bA_t$ as
\begin{align*}
\bA_t &= \bI + \lambda^{-1}\varphi(\bX)^\transp\varphi(\bX)
= \bI + \lambda^{-1}\sum_{s=1}^t\varphi(\bx_s)\varphi(\bx_s)^\transp\\
&= \bI +\lambda^{-1}\sum_{s=1}^t\varphi(\bx_{\fb{s}+1})\varphi(\bx_{\fb{s}+1})^\transp
= \bI + \lambda^{-1}\sum_{j=1}^h B_j\varphi(\bx_{j})\varphi(\bx_{j})^\transp\\
&= \bI + \lambda^{-1}\sum_{j=1}^h [\bW_h]_{j,j}\varphi(\bx_{j})\varphi(\bx_{j})^\transp
= \bI + \lambda^{-1}\varphi(\bX_{h})^\transp\bW_h\varphi(\bX_{h}),
\end{align*}
where $\bW_h$ is defined as described in \Cref{eq:mu-post-gp}. Applying
now steps $a-d$ in reverse,
with the only difference being the application of \Cref{eq:prim-dual-eq}
to $\bW_h^{1/2}\varphi(\bX_{h})$ rather than $\varphi(\bX_{t})$ in step $b$,
we obtain
\begin{align*}
{\sigma}_t^2(\bx_i)
&= \varphi(\bx_i)^\transp\bA_{t}^{-1}\varphi(\bx_i)
= \varphi(\bx_i)^\transp(\varphi(\bX_{h})^\transp\bW_h\varphi(\bX_{h})/\lambda + \bI)^{-1}\varphi(\bx_i)\\
&\stackrel{d,c}{=} \varphi(\bx_i)^\transp\varphi(\bx_i) - \varphi(\bx_i)^\transp\varphi(\bX_t)^\transp\bW_h^{1/2}\bW_h^{1/2}\varphi(\bX_t)(\varphi(\bX_t)^\transp\bW_h^{1/2}\bW_h^{1/2}\varphi(\bX_t) + \lambda\bI)^{-1}\varphi(\bx_i)\\
&\stackrel{b}{=} \varphi(\bx_i)^\transp\varphi(\bx_i) - \varphi(\bx_i)^\transp\varphi(\bX_t)^\transp\bW_h^{1/2}(\bW_h^{1/2}\varphi(\bX_t)\varphi(\bX_t)^\transp\bW_h^{1/2} + \lambda\bI)^{-1}\bW_h^{1/2}\varphi(\bX_t)\varphi(\bx_i)\\
&\stackrel{a}{=}
\kerfunc(\bx_i, \bx_i) - \kerfunc(\bx_i, \bX_t)\bW_h^{1/2}(\bW_h^{1/2}\bK_t\bW_h^{1/2} + \lambda\bI)^{-1}\bW_h^{1/2}\kerfunc(\bX_t, \bx_i)\\
&=
\kerfunc(\bx_i, \bx_i) - \kerfunc(\bx_i, \bX_t)(\bK_t + \lambda\bW_h^{-1})^{-1}\kerfunc(\bX_t, \bx_i),
\end{align*}
where in the last equality we simply collected $\bW_h$ to obtain the formulation
of \Cref{eq:mu-post-gp}. The reasoning for the mean is identical,
with one minor difference. After rewriting $\mu_t$ in its feature-space view,
and applying the fundamental equality
\begin{align*}
{\mu}_t(\bx_i) &= \kerfunc(\bx_i, \bX_t)(\bK_t + \lambda\bI)^{-1}\bK_t\by_t\\
&= \varphi(\bx_i)^\transp\varphi(\bX_t)^\transp(\varphi(\bX_t)\varphi(\bX_t)^\transp + \lambda\bI)^{-1}\varphi(\bX_t)\varphi(\bX_t)^\transp\by_t\\
&\stackrel{a}{=} \varphi(\bx_i)^\transp\varphi(\bX_t)^\transp\varphi(\bX_t)(\varphi(\bX_t)^\transp\varphi(\bX_t) + \lambda\bI)^{-1}\varphi(\bX_t)^\transp\by_t\\
&\stackrel{b}{=} \varphi(\bx_i)^\transp\varphi(\bX_h)^\transp\bW_h\varphi(\bX_h)(\varphi(\bX_h)^\transp\bW_h\varphi(\bX_h) + \lambda\bI)^{-1}\varphi(\bX_t)^\transp\by_t,
\end{align*}
where equality $a$ is once again due to \Cref{eq:prim-dual-eq},
and $b$ is due to the already proven equality
$\varphi(\bX_t)^\transp\varphi(\bX_t) = \varphi(\bX_h)^\transp\bW_h\varphi(\bX_h)$.
To handle the last remaining term $\varphi(\bX_t)^\transp\by_t$ we can rewrite
\begin{align*}
\varphi(\bX_t)^\transp\by_t
&= \sum_{s=1}^t\varphi(\bx_s)y_s
= \sum_{s=1}^t\varphi(\bx_{\fb{s}+1})y_{s}\\
&= \sum_{j=1}^h \varphi(\bx_{j})\sum_{s = t_j+1}^{t_j + B_h}y_s
= \sum_{j=1}^h \varphi(\bx_{j})[\by_h]_j
= \varphi(\bX_{h})^\transp\by_h,
\end{align*}
where once again $t_j$ is the step before the beginning of the $j$-th epoch,
that is $\fb{t} = t_j$ for all steps in the $j$-th epoch and the candidate
$\bx_{t_j + 1}$ is the one evaluated multiple times in the $j$-th epoch.
Putting it all together, and re-applying \Cref{eq:prim-dual-eq} we obtain
\begin{align*}
{\mu}_t(\bx_i) &= \varphi(\bx_i)^\transp\varphi(\bX_h)^\transp\bW_h\varphi(\bX_h)(\varphi(\bX_h)^\transp\bW_h\varphi(\bX_h) + \lambda\bI)^{-1}\varphi(\bX_h)^\transp\by_h\\
&= \varphi(\bx_i)^\transp\varphi(\bX_h)^\transp\bW_h^{1/2}(\bW_h^{1/2}\varphi(\bX_h)\varphi(\bX_h)^\transp\bW_h^{1/2} + \lambda\bI)^{-1}\bW_h^{1/2}\varphi(\bX_h)\varphi(\bX_h)^\transp\by_h\\
&= \kerfunc(\bx_i, \bX_h)(\bK_h + \lambda\bW_h^{-1})^{-1}\bK_h\by_h,
\end{align*}
which concludes the proof of the equivalence between \Cref{eq:post-gp}
and \Cref{eq:mu-post-gp}.

In this analysis we made two simplifications:
that the step $t$ was at the end of a batch, and that no two candidates
were the same in different batches.
To relax the first we can just consider extending $\bW_h$ to contain not only
all past $B_j$, but also a partial count of the current batch. Similarly
$\by_h$ has to be extended to include the partial feedback received during
the epoch. Similarly, if the same candidate was selected in two batches $j$
and $j'$, we simply have to merge their contributions in the sum
$\sum_{j=1}^h B_j\varphi(\bx_{j})\varphi(\bx_{j})^\transp$ e.g., by removing
the $j$-th term and account the $j'$-th term with $B_j + B_{j'}$ multiplicity.

\subsection{Proof of \Cref{lem:mu-switches}}
We begin with the following result from \cite{calandriello2020near}.
\begin{proposition}[{{Lem.~4} in \cite{calandriello2020near}}]\label{prop:ratio-bound-orig}
    For any kernel $\kerfunc$, set of points $\bX_t$, $\bx \in \armset$
    and $t < t'$
    \begin{align*}
        1 \leq \frac{\sigma_{t}^2(\bx)}{\sigma_{t'}^2(\bx)} \leq 1 + \sum_{s=t+1}^{t'}\sigma_{t}^2(\bx_s).
    \end{align*}
\end{proposition}
Using \Cref{prop:ratio-bound-orig} it is also very easy to show the following
known property of the one-step ratio between posteriors (i.e., $t' = t+1$)
\begin{align}\label{eq:one-step-ratio}
    1 \leq \frac{\sigma_{t}^2(\bx)}{\sigma_{t+1}^2(\bx)}
    \leq 1 + \sigma_{t}^2(\bx_{t+1})
\leq 1 + \sigma_{0}^2(\bx_{t+1})
= 1 + \kerfunc(\bx_{t+1}, \bx_{t+1})/\lambda \leq 1 + \kappa/\lambda,
\end{align}
where the first inequality is due to \Cref{prop:ratio-bound-orig},
the second due to the monotonicity of the posterior in $t$, and the
third due to our assumption $\kerfunc(\bx, \bx) \leq \kappa^2$.

Applying \Cref{prop:ratio-bound-orig} to our setting, where $\bx_s$ does not change for the whole batch,
and our epoch termination rule from \Cref{alg:meta-mini} we obtain
the following corollary
\begin{corollary}\label{cor:ratio-bound}
    During epoch $h$, let $t_h$ be the step before the beginning of the batch,
    and let
    $\bx_{t_h + 1}$ be the candidate selected for the whole batch at step $t_h + 1$.
    Then for any kernel $\kerfunc$, set of points $\bX_{t_h}$,
    $\bx \in \armset$,
    and $t_h + 1 = \fb{t} + 1 \leq t \leq \fb{t} + B_{h}$
    we have
    \begin{align*}
        1 \leq \frac{\sigma_{\fb{t}}^2(\bx)}{\sigma_t^2(\bx)} \leq 1 + B_h\sigma_{\fb{t}}^2(\bx_{\fb{t}+1}).
    \end{align*}
    Moreover, selecting $B_{\epoch} = \lfloor (C^2 - 1)/\sigma_{t_h}^2(\bx_{t_h+1}) \rfloor$
    we have $\frac{\sigma_{\fb{t}}(\bx)}{\sigma_t(\bx)} \leq C$.
\end{corollary}

To bound the number of epochs and prove \Cref{lem:mu-switches}
we follow a blueprint from
\cite{calandriello2020near}.
\begin{proof}[Proof of \Cref{lem:mu-switches}]
To bound the number of epoch we start from the fundamental inequality based
on the choice of $B_h$ and the properties of the floor function. Consider
an arbitrary $B_j$,
\begin{align*}
    &B_j \geq \frac{C^2 - 1}{\sigma_{t_j}^2(\bx_{t_j+1})} - 1\\
&\Rightarrow
\sigma_{t_j}^2(\bx_{t_j+1})(B_j + 1) \geq C^2 - 1\\
&\Rightarrow
2\sigma_{t_j}^2(\bx_{t_j+1})B_j \geq C^2 - 1.
\end{align*}
Note that due to the construction of the batch
$\sigma_{t_j}^2(\bx_{t_j+1}) = \sigma_{t_j}^2(\bx_{t_j+2}) = \dots = \sigma_{t_j}^2(\bx_{t_j+B_j})$,
and therefore $\sigma_{t_j}^2(\bx_{t_j+1})B_j = \sum_{s=t_j + 1}^{t_j + B_j}\sigma_{t_j}^2(\bx_{s})$.
Summing across batches up to batch $h$ we have
\begin{align*}
     h(C^2 - 1) &\leq 2\sum_{j=1}^h\sigma_{t_j}^2(\bx_{t_j+1})B_j
=
    2\sum_{j=1}^h\sum_{s=t_j + 1}^{t_j + B_j}\sigma_{t_j}^2(\bx_{s})\\
    &=2\sum_{j=1}^h\sum_{s=t_j + 1}^{t_j + B_j}\frac{\sigma_{t_j}^2(\bx_{s})}{\sigma_{s-1}^2(\bx_{s})}\sigma_{s-1}^2(\bx_{s})
    \stackrel{\text{\Cref{cor:ratio-bound}}}{\leq} 2\sum_{j=1}^h\sum_{s=t_j + 1}^{t_j + B_j}C^2\sigma_{s-1}^2(\bx_{s})\\
    &= 2C^2\sum_{s=1}^T\sigma_{s-1}^2(\bx_{s}).
\end{align*}
Now that we obtain the sum of posterior variances we need to connect this
quantity to $\gamma_T$. In order to do this we will use
a standard bound (see e.g., \cite{hazan_logarithmic_2006}) on the summation
\begin{align*}
    \sum_{s=1}^T\sigma_{s}^2(\bx_{s}) \leq \logdet(\bI + \kerfunc(\bX_t, \bX_t)/\lambda)
    = 2\gamma(\bX_t,\by_t) \leq 2\gamma_T.
\end{align*}
All that is left is to use \Cref{eq:one-step-ratio} to bound
$\sigma_{s-1}^2(\bx_{s})$ in terms of $\sigma_{s}^2(\bx_{s})$ and rearrange
appropriately all the derived results to obtain
\begin{align*}
    h &\leq \frac{2C^2}{C^2 - 1}\sum_{s=1}^T\sigma_{s-1}^2(\bx_{s})
    \leq \frac{2C^2}{C^2 - 1}(1 + \kappa^2/\lambda)\sum_{s=1}^T\sigma_{s}^2(\bx_{s})
    \leq \frac{4C^2}{C^2 - 1}(1 + \kappa^2/\lambda)\gamma_T
    \leq \bigotime\left(\frac{C^2}{C^2 - 1}\gamma_T\right)
\end{align*}
\end{proof}

\subsection{Proof of \Cref{lem:instaregbound}}
The proof of this result follows directly from the sketch of proof in
\Cref{sec:meta-regret}, combined with \Cref{cor:ratio-bound} here in the appendix.

\subsection{Proofs of \Cref{sec:meta-apply}}
We indicate here how each original regret proofs can be modified to obtain
\Cref{thm:mu-gp-ucb-freq,thm:mu-gp-ucb-bayes,thm:mu-gp-ei-freq}.
All proof will depend on a standard blueprint \cite{srinivas2010gaussian,abbasi2011improved,chowdhury2017kernelized,calandriello2020near} that we present first.
For simplicity we also assume again that step $T$ is exactly at the end of batch
$h$. This is without loss of generality as we can always artificially truncate
the current batch at step $T$ for the sake of the analysis without increasing
the regret. (i.e., \Cref{cor:ratio-bound} and all other results will still
hold).

First we leverage results from the original analyses to show
that the instantaneous regret $r_{t_j + 1}$ at the beginning of
batch $j$ is bounded (i.e., \Cref{eq:instant-regret}) as
\begin{align*}
    r_{t_j + 1} \leq G_{t_j}\sigma_{t_{j}}(\bx_{t_j + 1}).
\end{align*}
Moreover, the same candidate is selected at all steps in the batch, so it holds
as well that for all $t' \in [t_j+1, t_j + B_j]$ we have
\begin{align*}
r_{t'}
    = r_{t_j + 1} \leq G_{t_j}\sigma_{t_{j}}(\bx_{t_j + 1})
    = G_{t_j}\sigma_{t_{j}}(\bx_{t'})
\end{align*}
Then, leveraging again \Cref{cor:ratio-bound} to bound the posterior ratios we obtain
\begin{align*}
    \sum_{s=t_j+1}^{t_j+ B_j}
    r_{s} \leq G_{t_j}\sum_{s=t_j+1}^{t_j+ B_j}\sigma_{t_{j}}(\bx_{s})
    \stackrel{\text{\Cref{cor:ratio-bound}}}{\leq} G_{t_j}C\sum_{s=t_j+1}^{t_j+ B_j}\sigma_{s-1}(\bx_{s}).
\end{align*}
Finally, using the fact that $G_t$ is non-decreasing, and summing across batches
\begin{align*}
    R_T &= \sum_{t=1}^T r_t
    = \sum_{j=1}^h \sum_{s=t_j+1}^{t_j+ B_j} r_{s}\\
    &\leq \sum_{j=1}^h G_{t_j}\sum_{s=t_j+1}^{t_j+ B_j}\sigma_{t_{j}}(\bx_{s})
    \leq \sum_{j=1}^h G_{t_j}C\sum_{s=t_j+1}^{t_j+ B_j}\sigma_{s-1}(\bx_{s})\\
    &\leq G_T C \sum_{j=1}^h \sum_{s=t_j+1}^{t_j+ B_j}\sigma_{s-1}(\bx_{s})
    \leq G_T C \sum_{s=1}^T \sigma_{s-1}(\bx_{s}).
\end{align*}
Putting it all together and expressing everything in terms of $\gamma_T$ we derive
\begin{align*}
R_T &\leq G_T C \sum_{s=1}^T \sigma_{s-1}(\bx_{s})\\
&\stackrel{a}{\leq} G_T C \sqrt{T}\sqrt{\sum_{s=1}^T \sigma_{s-1}^2(\bx_{s})}\\
&\stackrel{b}{\leq} G_T C \sqrt{T}\sqrt{\left(1 + \tfrac{\kappa^2}{\lambda}\right)\sum_{s=1}^T \sigma_{s}^2(\bx_{s})}\\
&\stackrel{c}{\leq} G_T C \sqrt{T}\sqrt{\left(1 + \tfrac{\kappa^2}{\lambda}\right)\gamma_T},
\end{align*}
where $a$ is due to Cauchy-Schwarz inequality,
$b$ due to \Cref{eq:one-step-ratio}, and $c$ due to the usual bounding
of posterior variances with the $\logdet$ and information gain \cite{hazan_logarithmic_2006}.
Now that we have this blueprint, all that is left is to look at results in the
literature on how $G_t$ can be bounded under different assumptions and acquisition
functions.

\begin{proof}[Proof of \Cref{thm:mu-gp-ucb-freq}]
For this theorem we leverage the assumptions $\normsmall{f} \leq \fnorm$
with $u_{t}^{\text{\mugpucb}}$ as acquisition function. Already Theorem 1
from \cite{chowdhury2017kernelized} showed that if $\lambda = \xi^2$ and
$\beta_t$
is tuned as $\beta_t = \Theta(\sqrt{\logdet(\bK_t/\xi^2 + \bI) + \log(1/\delta)} + F)$
we obtain that $G_t \leq \beta_t$ suffices to guarantee \Cref{eq:instant-regret}.
Note however that in our case we can efficiently compute
$\logdet(\bK_t/\xi^2 + \bI)$ leveraging the few unique candidates. In particular,
since we only need to compute $\beta_{t_h}$ at the beginning of a batch,
we can rewrite
\begin{align*}
    \logdet(\bK_t/\xi^2 + \bI)
    &= \logdet(\varphi(\bX_t)\varphi(\bX_t)^\transp/\xi^2 + \bI)\\
    &\stackrel{a}{=} \logdet(\varphi(\bX_t)^\transp\varphi(\bX_t)/\xi^2 + \bI)\\
    &\stackrel{b}{=} \logdet(\varphi(\bX_h)^\transp\bW_h\varphi(\bX_h)/\xi^2 + \bI)\\
    &\stackrel{c}{=} \logdet(\bW_h^{1/2}\varphi(\bX_h)^\transp\varphi(\bX_h)\bW_h^{1/2}/\xi^2 + \bI)\\
    &= \logdet(\bW_h^{1/2}\bK_h\bW_h^{1/2}/\xi^2 + \bI),
\end{align*}
where $a$ is due to Sylvester's determinant identity, $b$ is our usual re-writing,
and $c$ is Sylvester's determinant identity again.
Not that since this a strict equality, we still have that at the last batch $h$
\begin{align*}
\logdet(\bW_h^{1/2}\bK_h\bW_h^{1/2}/\xi^2 + \bI)
= \logdet(\bK_T/\xi^2 + \bI) \leq \gamma_T,
\end{align*}
which gives the second half of the theorem.
\end{proof}
\begin{proof}[Proof of \Cref{thm:mu-gp-ucb-bayes}]
    This is a direct consequnce of Theorem 1 from \cite{srinivas2010gaussian}.
    In particular, they once again show that $G_t \leq \beta_t$ suffices
    to guarantee \Cref{eq:instant-regret}.
\end{proof}
\begin{proof}[Proof of \Cref{thm:mu-gp-ei-freq}]
    For this combination our starting point is Equation (38) in \cite{wang2014theoretical},
    which states that for
    \begin{align*}
\nu_t = 
\sqrt{\logdet(\bK_t/\lambda + \bI) + \sqrt{\logdet(\bK_t/\lambda + \bI)\log(t/\delta)} +\log(t/\delta)},
&& \lambda = \xi^2,
    \end{align*}
running a standard \gpopt loop (i.e., \Cref{alg:opt-gp}) with $u_t = u_t^{\text{\gpei}}$
guarantees
\begin{align*}
    r_t \leq \abigotime\left(  \left(\sqrt{F^2 + \gamma_t} + \nu_t\right)\sigma_{t-1}(\bx_t)\right),
\end{align*}
where we greatly simplified their notation by ignoring constant and logarithmic
terms. Noticing now that at each batch $j$ the values of $\beta_j$ and
$\nu_{t_j}$ are equal thanks again to the equivalence
$\logdet(\bK_t/\lambda + \bI) = \logdet(\bW_h^{1/2}\bK_h\bW_h^{1/2}/\lambda + \bI)$,
we can obtain a bound on $G_t$ that satisfies \Cref{eq:instant-regret} as
\begin{align*}
G_t \leq \abigotime\left(  \left(\sqrt{F^2 + \gamma_t} + \beta_t\right)\right)
\leq \abigotime\left(  \left(\sqrt{F^2 + \gamma_t} + \sqrt{\gamma_t}\right)\right)
\leq \abigotime\left(  \left(F + \sqrt{\gamma_t}\right)\right).
\end{align*}
Putting this together with the usual blueprint to bound the error incurred by
batching and the bound on the sum of posterior variances with $\gamma_T$
we obtain our result.
\end{proof}
\section{Extended experimental results}
\begin{table}[t]
\centering
\begin{tabular}{ll}
    \mugpucb & $\sigma = 455.56$, $C = 1.1$\\
    \mugpei & $\sigma = 455.56$, $C = 1.1$\\
\bbkb & $\sigma = 277.78$, $C = 1.1$\\
\bkb & $\sigma = 455.56$\\
\gpucb & $\sigma = 500.00$\\
\bgpucb & $\sigma = 455.56$, $C = 1.1$\\
$\varepsilon$-\textsc{greedy} & $a = 1$, $b = 0.5$
\end{tabular}
\caption{Optimal hyper-parameters found for the NAS-bench experiment.}\label{tab:hyperparm}
\end{table}
\begin{figure}[t]
    \centering
     \small{\vcenteredinclude{legend.pdf}}

     \vspace{.5\baselineskip}
	 \begin{subfigure}[b]{0.49\textwidth}
		 \centering
         \includegraphics[width=\textwidth,trim=.2cm .2cm 4cm .2cm,clip]{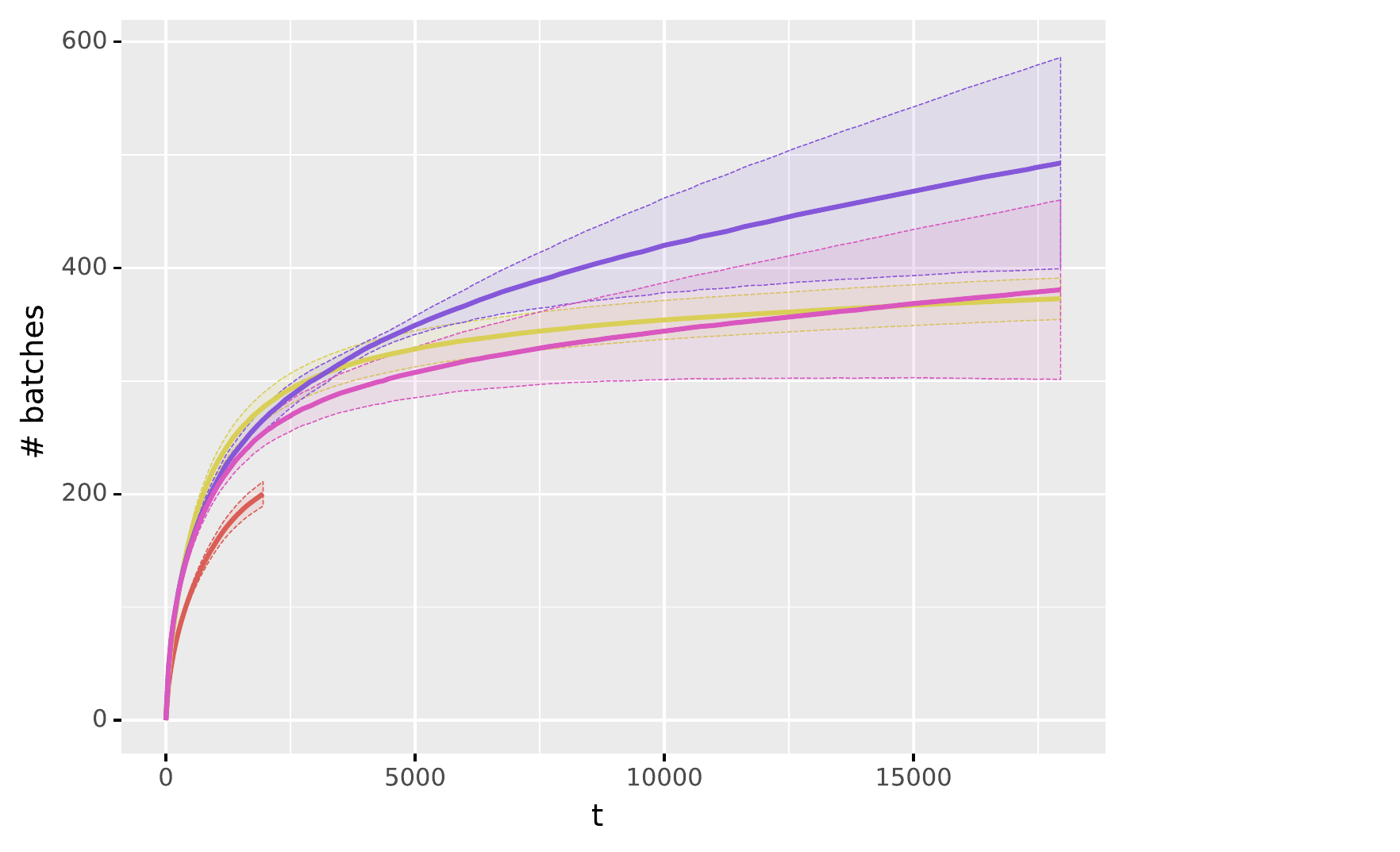}
		 \caption{Batch/epoch number $h$ against step $t$}
		 \label{fig:batch}
	 \end{subfigure}
	 \hfill
	 \begin{subfigure}[b]{0.49\textwidth}
		 \centering
	\includegraphics[width=\textwidth,trim=.2cm .2cm 4cm .2cm,clip]{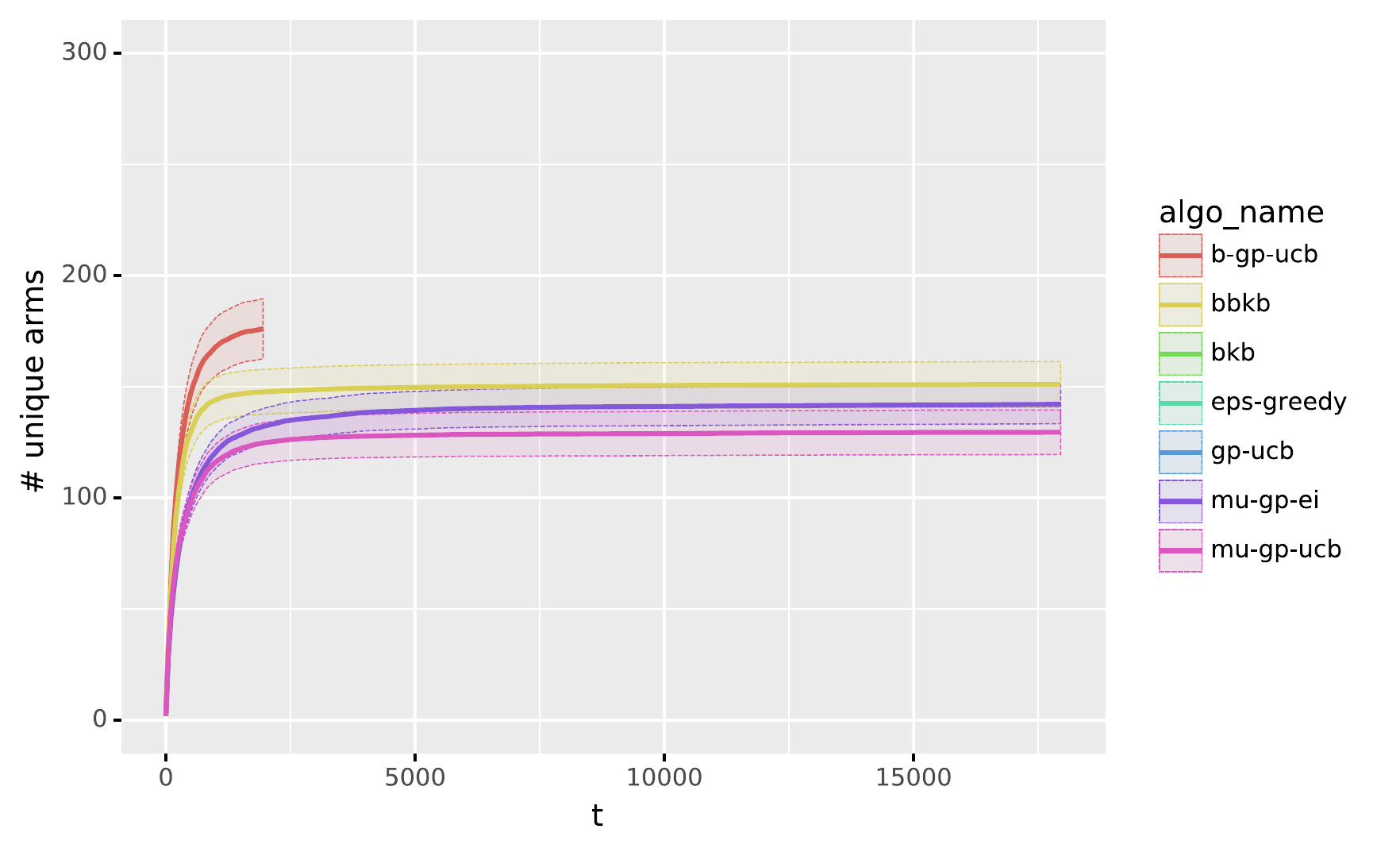}
		 \caption{$\uq_t$ against step $t$ }
		 \label{fig:unique-mod}
	 \end{subfigure}
	 \caption{}
\end{figure}
We report here additional details on the NAS-bench experiments, as well
as an evaluation on synthetic functions.

\subsection{Additional details on NAS-bench experiments}
All the implementations are based on the code released by \cite{calandriello2020near}
for \bbkb and \bkb, available at \url{https://github.com/luigicarratino/batch-bkb}.
For each algorithm we run the experiment 40 times with different seeds,
and select the hyper-parameters that achieves the lowest average regret.
Hyper-parameters for all algorithms based on GPs are searched among
\begin{align*}
    \sigma^2 &= \{100.00, 144.45, 188.89, 233.33, 277.78, 322.22, 366.67, 411.11, 455.56, 500.00\}\\
    C &= \{1.1, 1.2\}
\end{align*}
where $C$ is the threshold used for batching, and $\sigma$ is the bandwidth
of the Gaussian kernel.
For $\varepsilon$-\textsc{greedy} the exploration rate is
is optimized as $\varepsilon = a/t^b$ over $a \in \{0.1, 1, 10\}$ and $b \in \{1/3, 1/2, 1, 2\}$.
The empirically optimal values are reported in \Cref{tab:hyperparm}.

Beyond regret, runtime and unique candidates reported in \Cref{sec:exp}, another interesting figure to empirically measure is the difference between the number $h$
of batches/switches at step $T$ and the number of unique candidates $\uq_T$
selected. As we know, $h$ is an upper bound for $\uq_T$, but looking
at \Cref{fig:batch} and \Cref{fig:unique-mod} we see that especially in the later
stages of the optimization these two quantities have a significant gap between them.
This can be explained by noticing that as the optimization progresses the
algorithm tend to focus on a few good candidates, switching back and forth between
them which increases $h$ but not $\uq_T$.

\subsection{Evaluation on synthetic functions}
\begin{figure}[t]
    \centering
     \small{\vcenteredinclude{legend.pdf}}

     \vspace{.5\baselineskip}
	 \begin{subfigure}[b]{0.41\textwidth}
		 \centering
         \includegraphics[width=\textwidth,trim=.2cm .2cm 4cm .2cm,clip]{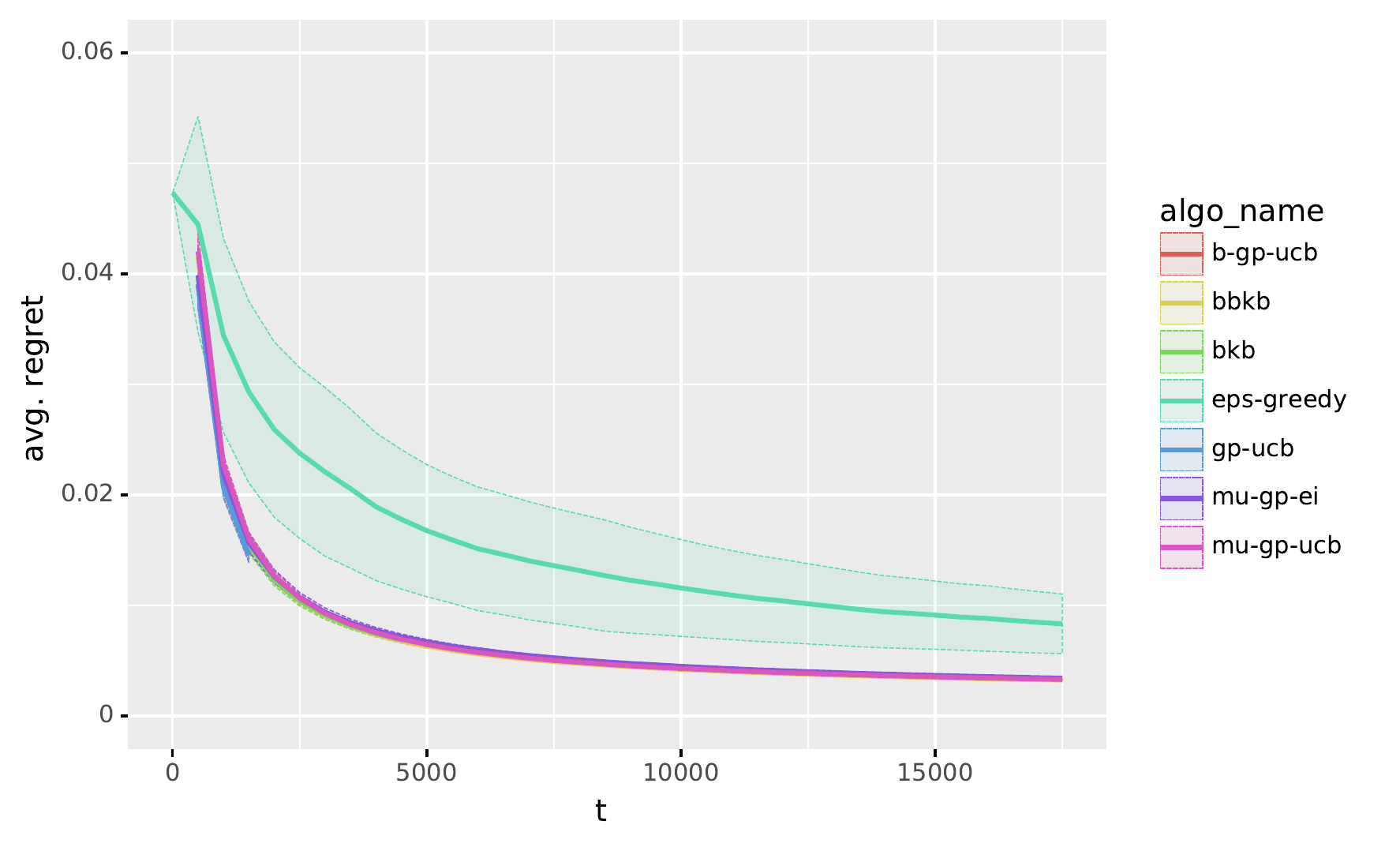}
		 \caption{Ellipsoid}
		 \label{fig:ellipsoid}
	 \end{subfigure}
	 \hfill
	 \begin{subfigure}[b]{0.41\textwidth}
		 \centering
	\includegraphics[width=\textwidth,trim=.2cm .2cm 4cm .2cm,clip]{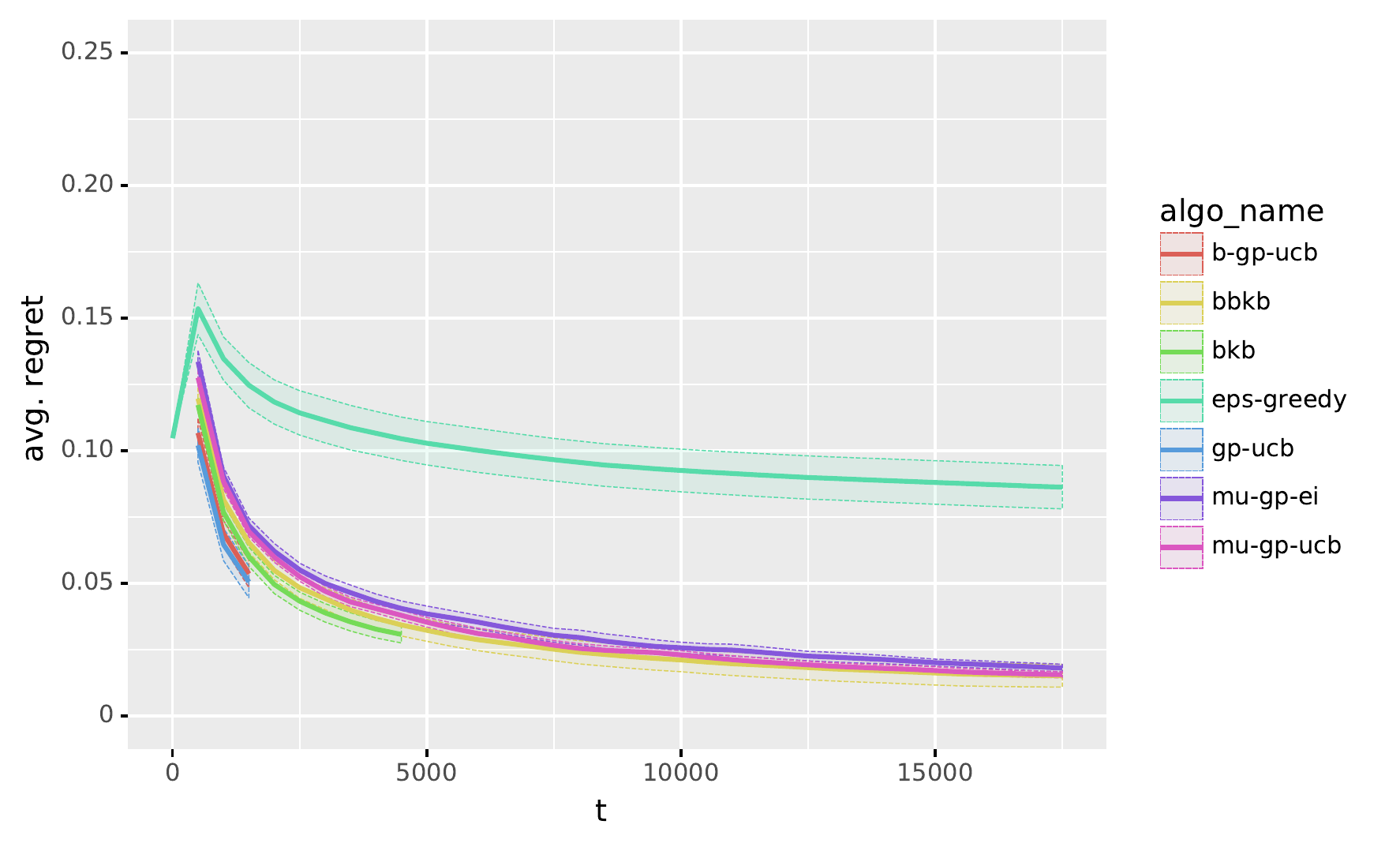}
		 \caption{Rastigrin}
		 \label{fig:rastigrin}
	 \end{subfigure}

	 \begin{subfigure}[b]{0.41\textwidth}
		 \centering
         \includegraphics[width=\textwidth,trim=.2cm .2cm 4cm .2cm,clip]{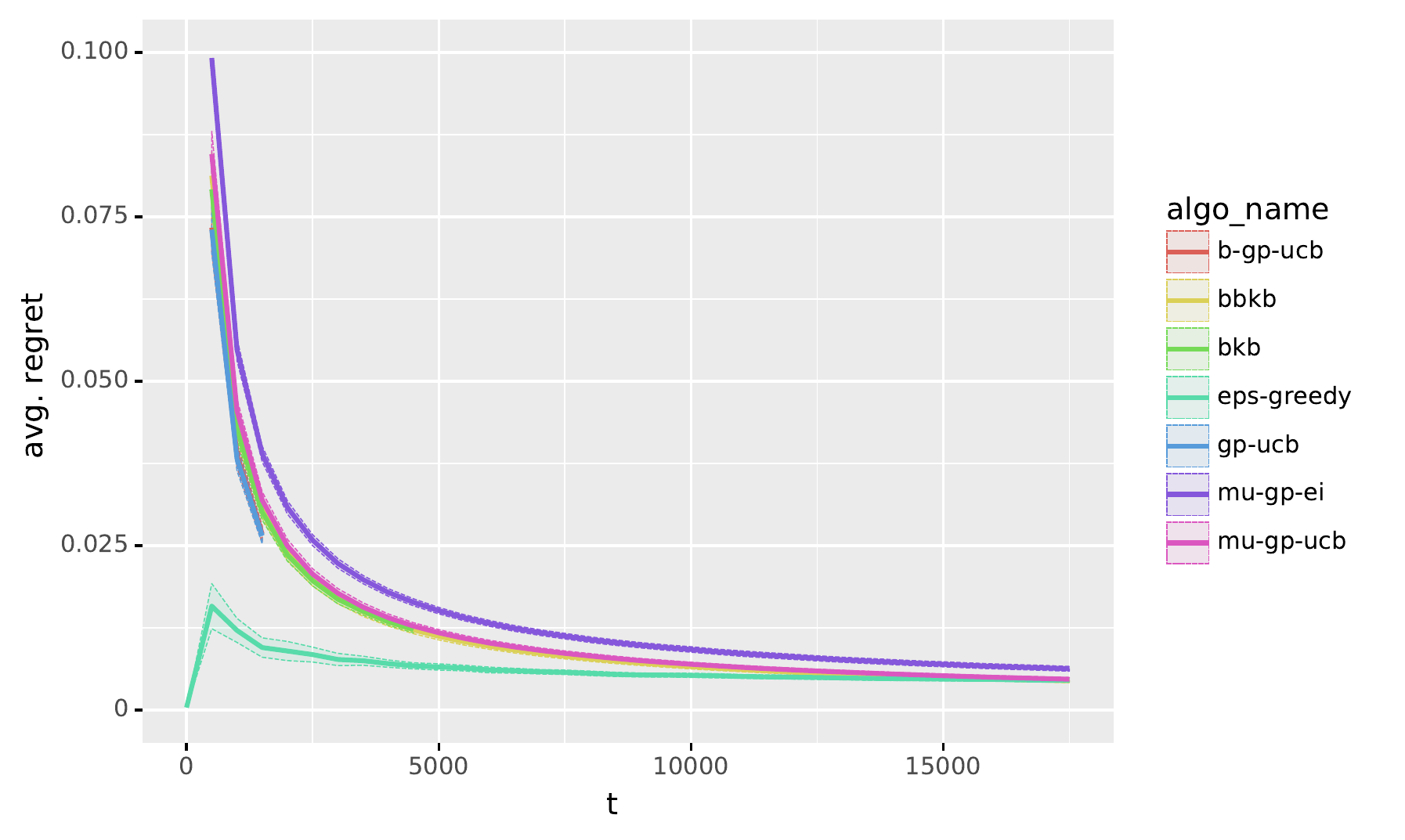}
		 \caption{Rosenbrock}
		 \label{fig:rosenbrock}
	 \end{subfigure}
	 \hfill
	 \begin{subfigure}[b]{0.41\textwidth}
		 \centering
	\includegraphics[width=\textwidth,trim=.2cm .2cm 4cm .2cm,clip]{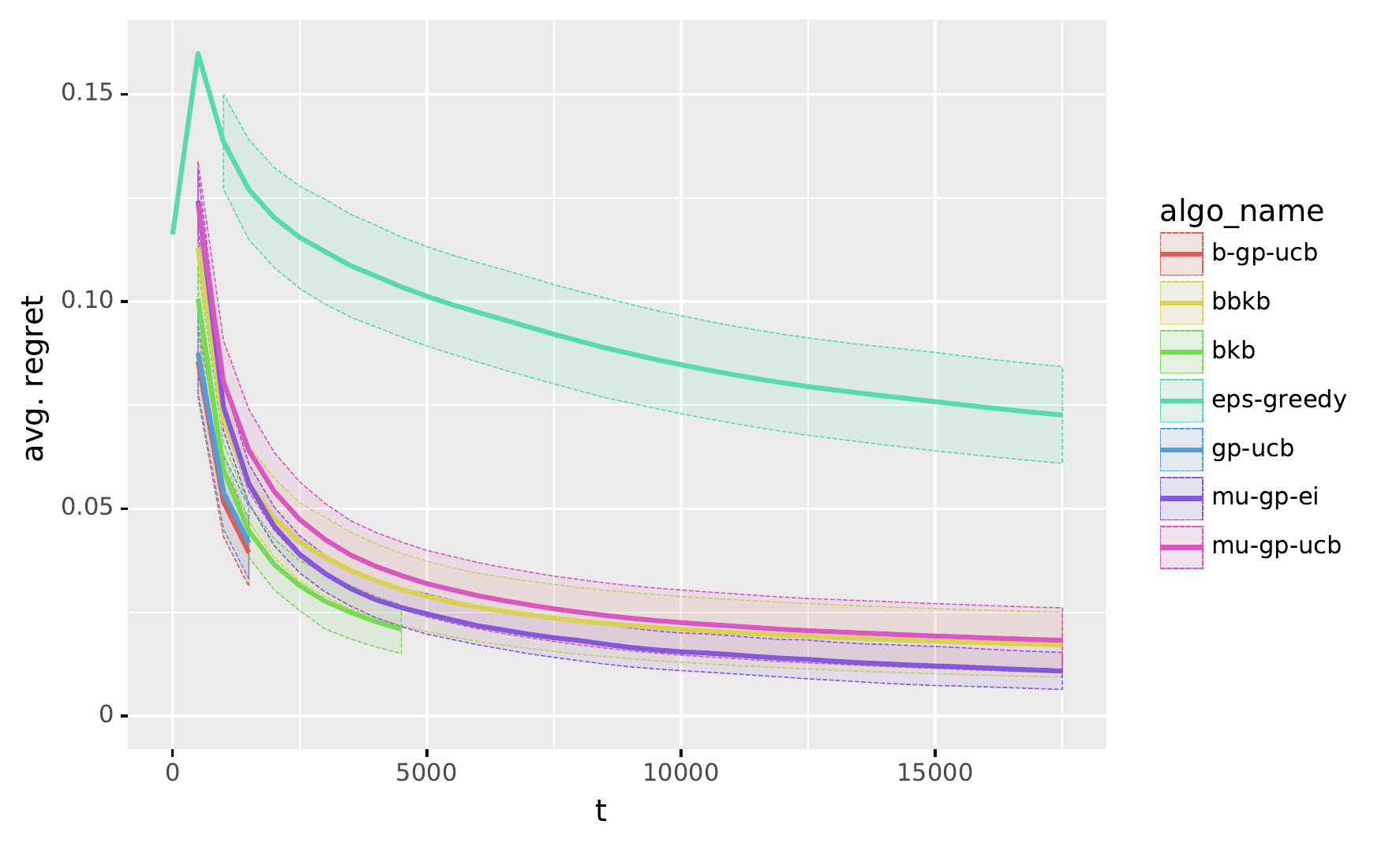}
		 \caption{Schaffer}
		 \label{fig:schaffer}
	 \end{subfigure}
     \caption{$R_t/t$ against step $t$}
\end{figure}
We also evaluate our method on common benchmark functions from the noisy
BBOB benchmark suite \cite{hansen2010real}.
In particular we focus on the Rosenbrock ($f_{104}$), ellipsoid ($f_{116}$),
and Schaffer ($f_{122}$) functions, all with moderate Gaussian noise.
We also include an extra function, the separable Rastigrin function, to
add another more complex separable function with the ellipsoid.

All functions are defined on $\Real^3$, with each coordinate
split into 21 sections with the same length by 22 evenly spaced points
placed between $[-5, 5]$.
The resulting discrete grid of points represent our candidate set
and contains $|\armset| = 22^3 = 10648$ unique candidates.
Similarly to the NAS-bench experiments, $\sigma$ of the Gaussian kernel,
$C$ and the $\varepsilon$-\textsc{GREEDY} parameters are selected optimally
over a grid search.

\begin{figure}[H]
    \centering
     \small{\vcenteredinclude{legend.pdf}}

     \vspace{.5\baselineskip}
	 \begin{subfigure}[b]{0.41\textwidth}
		 \centering
         \includegraphics[width=\textwidth,trim=.2cm .2cm 4cm .2cm,clip]{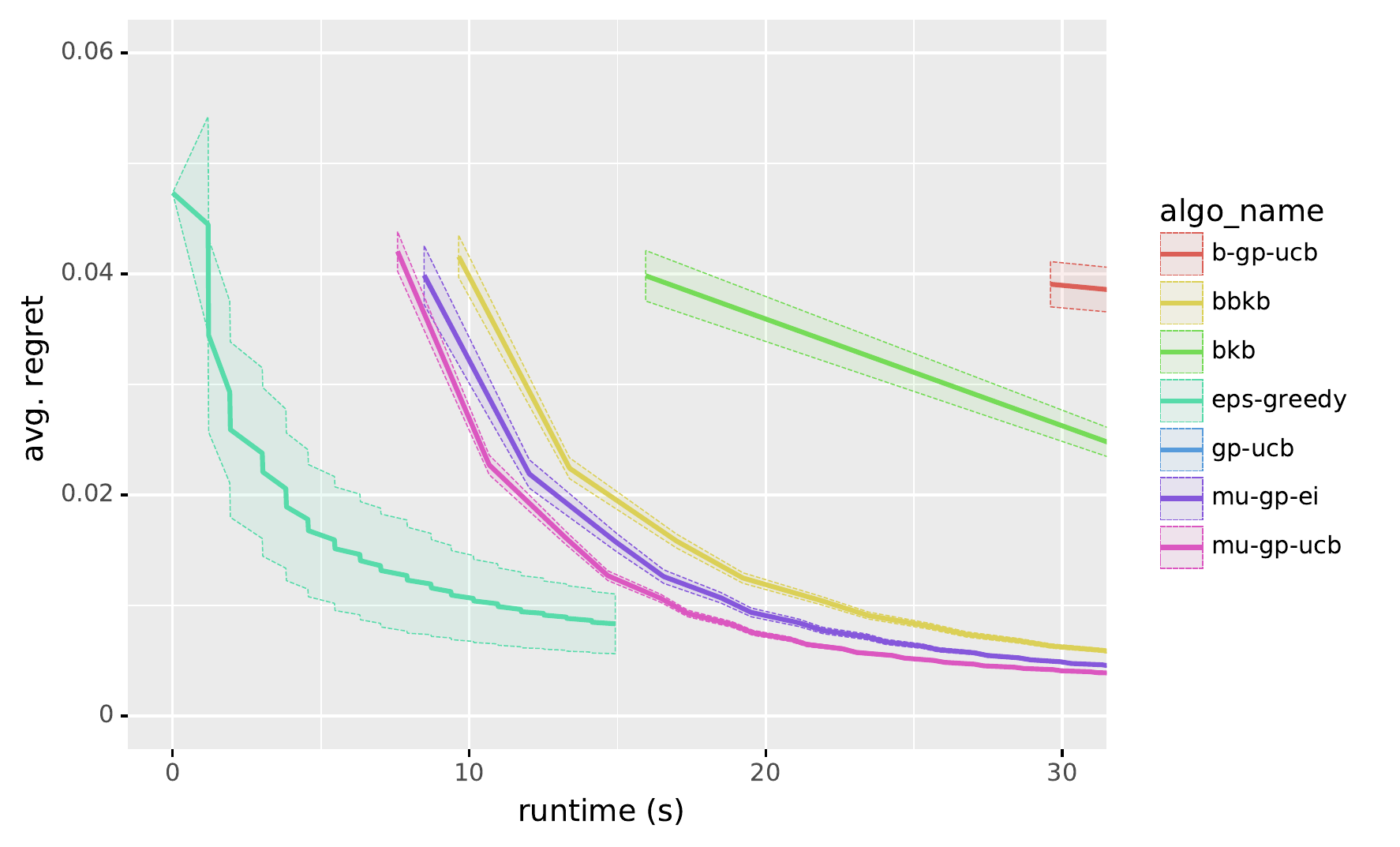}
		 \caption{Ellipsoid}
		 \label{fig:ellipsoid-t}
	 \end{subfigure}
	 \hfill
	 \begin{subfigure}[b]{0.41\textwidth}
		 \centering
	\includegraphics[width=\textwidth,trim=.2cm .2cm 4cm .2cm,clip]{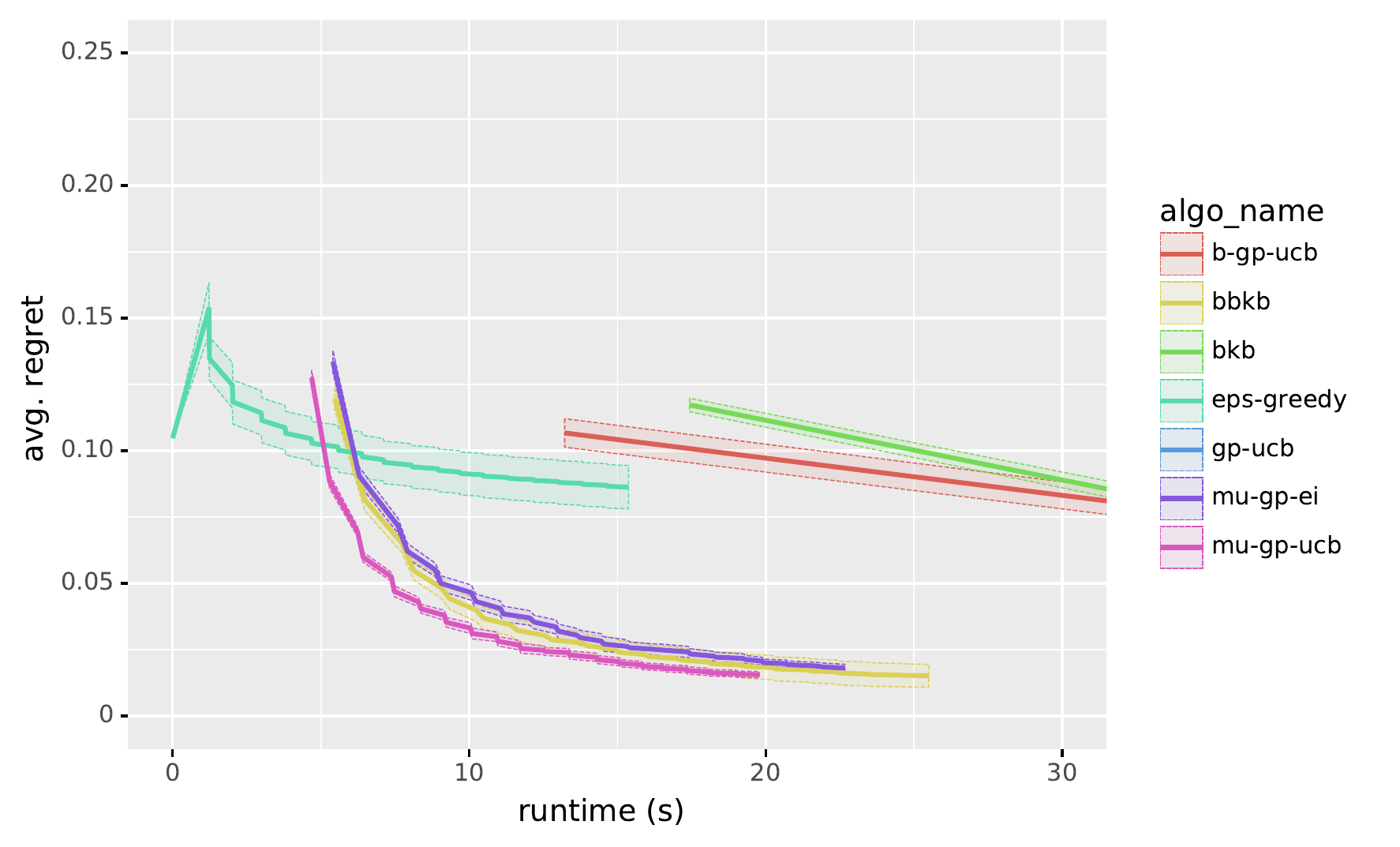}
		 \caption{Rastigrin}
		 \label{fig:rastigrin-t}
	 \end{subfigure}

	 \begin{subfigure}[b]{0.41\textwidth}
		 \centering
         \includegraphics[width=\textwidth,trim=.2cm .2cm 4cm .2cm,clip]{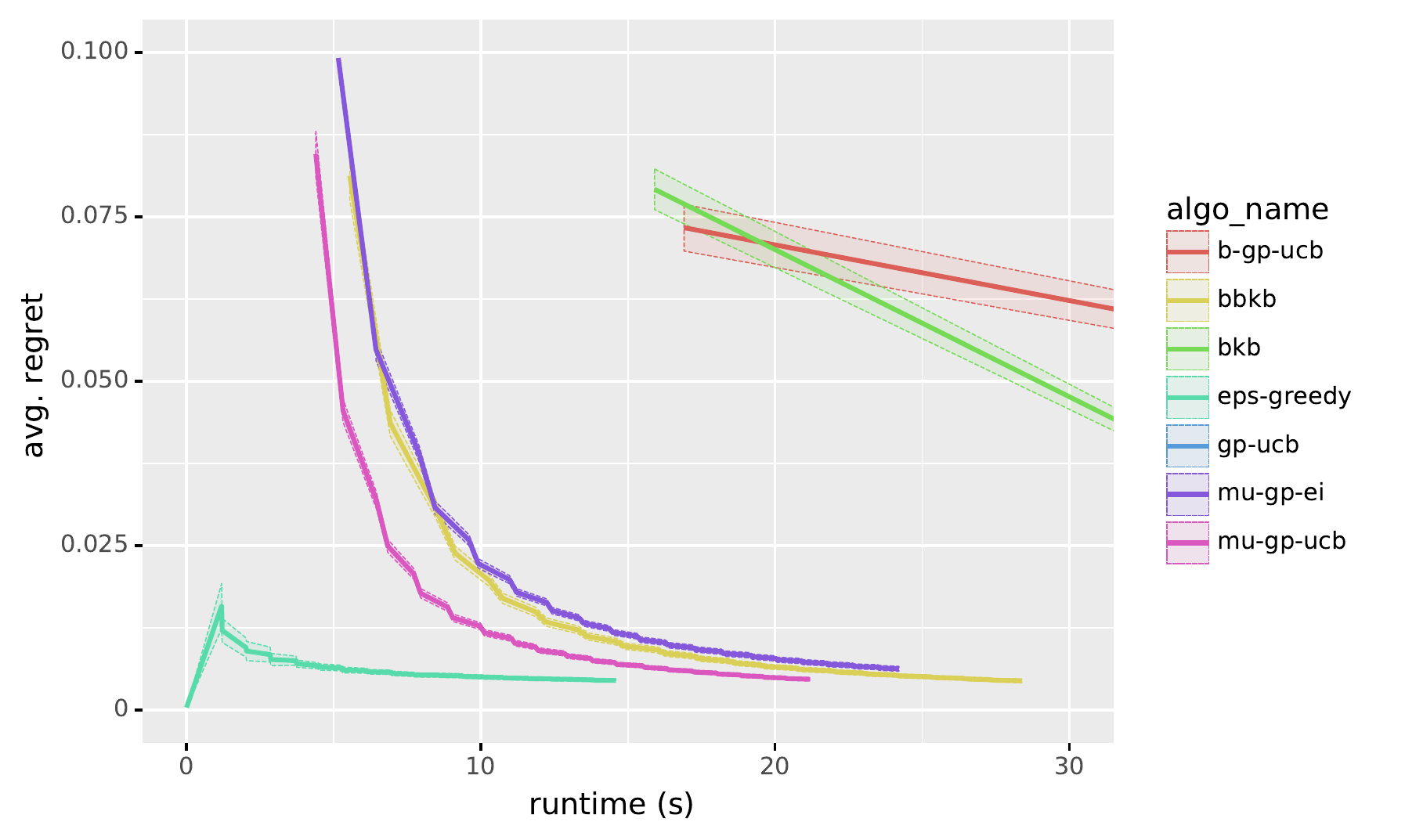}
		 \caption{Rosenbrock}
		 \label{fig:rosenbrock-t}
	 \end{subfigure}
	 \hfill
	 \begin{subfigure}[b]{0.41\textwidth}
		 \centering
	\includegraphics[width=\textwidth,trim=.2cm .2cm 4cm .2cm,clip]{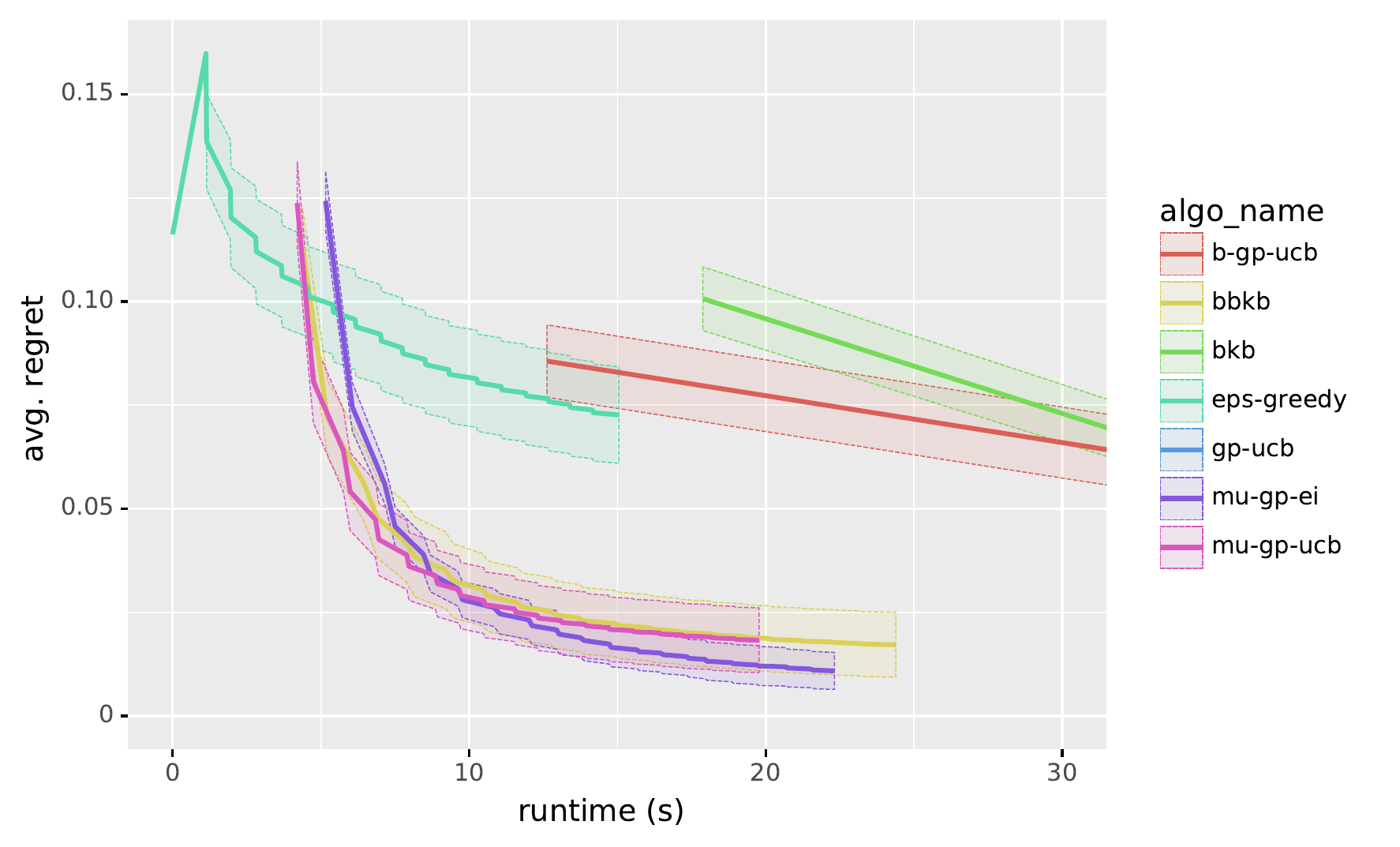}
		 \caption{Schaffer}
		 \label{fig:schaffer-t}
	 \end{subfigure}
     \caption{$R_t/t$ against time (in seconds)}
\end{figure}

The results on regret against steps are reported in 
\Cref{fig:ellipsoid,fig:rastigrin,fig:rosenbrock,fig:schaffer}
and on regret against time in 
\Cref{fig:ellipsoid-t,fig:rastigrin-t,fig:rosenbrock-t,fig:schaffer-t}.
As we can see, our approach achieve comparable runtime and regret to other
state of the art \gpopt methods. However, on more complex problem like the
Rosenbreck function, even the flexibility of a GP is not capable of capturing
the underlying shape of the optimization problem, and all \gpopt methods,
including ours, only perform roughly as well as a tuned $\varepsilon$-\textsc{GREEDY}
exploration.

\end{document}